\documentclass[10pt,twocolumn,letterpaper]{article}

\usepackage[pagenumbers]{cvpr} % To force page numbers, e.g. for an arXiv version

\definecolor{cvprblue}{rgb}{0.21,0.49,0.74}
\usepackage[pagebackref,breaklinks,colorlinks,allcolors=cvprblue]{hyperref}

\usepackage{algorithm,algpseudocode}
\usepackage{algorithmicx}

\usepackage{graphicx}
\usepackage{amssymb}
\usepackage{booktabs}

\usepackage[utf8]{inputenc} % allow utf-8 input
\usepackage[T1]{fontenc}    % use 8-bit T1 fonts
\usepackage{url}            % simple URL typesetting†
\usepackage{booktabs}       % professional-quality tables
\usepackage{amsfonts}       % blackboard math symbols
\usepackage{nicefrac}       % compact symbols for 1/2, etc.
\usepackage{microtype}      % microtypography
\usepackage{xcolor}         % colors

\usepackage{color}
\usepackage{amsmath}
\usepackage{amssymb}
\usepackage{mathtools}
\usepackage{arydshln}
\usepackage{amsthm}
\usepackage{bm}
\usepackage{enumitem}
\usepackage{multirow}
\usepackage{cuted}

\usepackage{tikz}

\newcommand{\srch}{\mathbf{x}^{\mathrm{high}}}
\newcommand{\srcl}{\mathbf{x}^{\mathrm{low}}}

\newcommand{\refl}{\mathbf{y}^{\mathrm{low}}}
\newcommand{\tgth}{\mathbf{y}^{\mathrm{high}}}
\newcommand{\imglr}{I^{\mathrm{low}}_\mathrm{ref}}
\newcommand{\imghr}{I^{\mathrm{high}}_\mathrm{ref}}
\newcommand{\imgls}{I^{\mathrm{low}}_\mathrm{src}}
\newcommand{\imghs}{I^{\mathrm{high}}_\mathrm{src}}
\newcommand{\methodname}{ScaleEdit}
\newcommand{\vdashline}{%
    \begin{tikzpicture}
        \draw[dashed, line width=0.5pt] (0,0) -- (0,2.5); 
    \end{tikzpicture}
}

\def\paperID{*****} % *** Enter the Paper ID here
\def\confName{CVPR}
\def\confYear{2026}

\title{Low-Resolution Editing is All You Need for High-Resolution Editing}

\author{
Junsung Lee\textsuperscript{\normalfont 1}\thanks{ Equal contribution.} \hspace{10mm} 
Hyunsoo Lee\textsuperscript{\normalfont 1}\footnotemark[1] \hspace{10mm} 
Yong Jae Lee\textsuperscript{\normalfont 3} \hspace{10mm} 
Bohyung Han\textsuperscript{\normalfont 1,2} \\ 
   \textsuperscript{1}ECE \&  \textsuperscript{2}IPAI, Seoul National University \hspace{5  mm}  \textsuperscript{3}University of Wisconsin-Madison \hspace{10mm} \\
   {\tt\small \{leejs0525,\,philip21,\,bhhan\}@snu.ac.kr, yongjaelee@cs.wisc.edu}
}

\begin{document}
\maketitle
% !TEX root = ./../main.tex

\begin{strip}
\centering
\vspace{-12mm}
\includegraphics[width=1.0\linewidth]{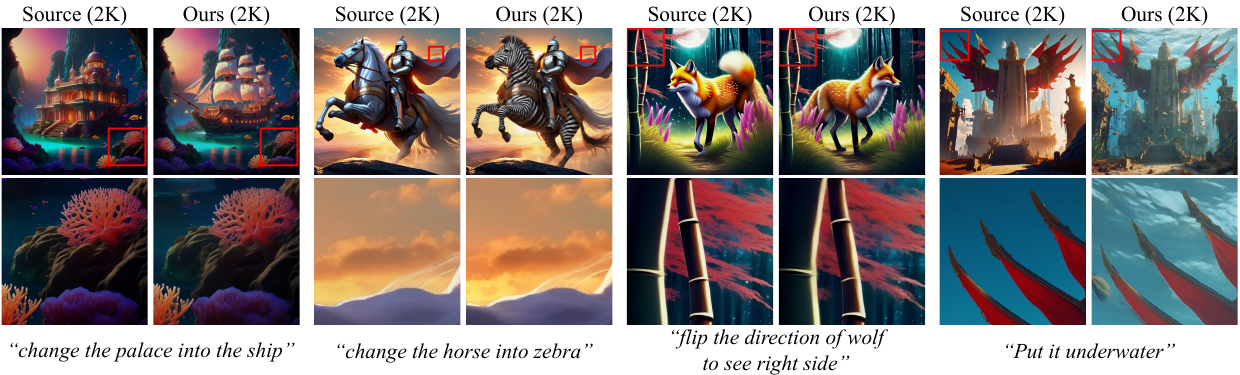}
\vspace{-7mm}
\captionof{figure}{
Image editing results of the proposed method, \methodname.
Our method successfully generates high-resolution edited images by leveraging low-resolution editing results as reference images.
}
\label{fig:teaser}
\end{strip}
\begin{abstract}
    High-resolution content creation is rapidly emerging as a central challenge in both the vision and graphics communities. 
    Images serve as the most fundamental modality for visual expression, and content generation that aligns with the user intent requires effective, controllable high-resolution image manipulation mechanisms.
    However, existing approaches remain limited to low-resolution settings, typically supporting only up to 1K resolution.
    In this work, we introduce the task of high-resolution image editing and propose a test-time optimization framework to address it.
    Our method performs patch-wise optimization on high-resolution source images, followed by a fine-grained detail transfer module and a novel synchronization strategy to maintain consistency across patches.
    Extensive experiments show that our method produces high-quality edits, facilitating high-resolution content creation.
    Project website: \url{https://hleephilip.github.io/ScaleEdit}
\end{abstract}

% !TEX root = ./../main.tex

\vspace{-5mm}
\section{Introduction}
\label{sec:intro}

High-resolution visual content has become essential in modern digital workflows, including image synthesis~\cite{du2024demofusion, huang2024fouriscale, qiu2025freescale, he2023scalecrafter, zhang2025diffusion},  video generation~\cite{blattmann2023align, hu2025ultragen, chen2025dc}, document processing~\cite{niu2025mineru2}, autonomous driving~\cite{wu2025drivescape}, and 3D shape generation~\cite{hunyuan3d22025tencent, lai2025hunyuan3d25highfidelity3d}.
In such scenarios, users often seek precise and intentional modifications to an existing high-resolution image rather than generating a new one from scratch.
Unlike na\"ive image generation, image manipulation preserves the identity and semantic layout of the source image, while allowing controlled adjustments that reflect the user’s intention.
Therefore, editing serves as a reliable strategy for producing user-controlled high-resolution content while preserving both structural integrity and visual fidelity.
This motivates the need for editing methods that maintain fine-grained texture details and structural consistency at high resolutions (\textit{e.g.}, > $\mathrm{1K} \approx 1024^2$).

We introduce the task of \emph{high-resolution image editing}, where the objective is to modify a high-resolution source image according to a target visual concept while preserving its fine-grained details.
However, existing image-to-image translation methods~\cite{comanici2025gemini, liu2025step1x-edit, zhu2025kv} operate at fixed low resolutions and cannot directly handle large-scale inputs. 
A na\"ive solution is to perform editing at low resolution and subsequently apply super-resolution methods~\cite{duan2025dit4sr, cheng2025effective, sun2024pisasr, dong2025tsd}, but this approach fails to recover micro-scale textures since the fine-grained details of the source image are not conditioned during the editing process.
Thus, a novel strategy is required to support editing in high-resolution.

To address the task defined in our work, we propose \emph{\methodname}.
Editing a high resolution image requires an optimization that preserves both semantics and fine-grained details. 
This can be facilitated by leveraging the strong performance of existing low-resolution image editing methods~\cite{liu2025step1x-edit, comanici2025gemini, zhu2025kv}.
Under this view, the central challenge of the task becomes clear: 
how to faithfully transfer the fine-scale details of the high-resolution source image into the edited result.
Our intuition is that this objective can be achieved by exploiting the generative priors.
Specifically, we constrain optimization to the sampling process of the pretrained generative models~\cite{rombach2022high, podell2023sdxl, flux2024}, implicitly enforcing the generative prior as a structural constraint that ensures plausible results.

Concretely, our key idea is to transfer fine-grained details from the high-resolution source image to the target image by introducing a learnable transfer function defined in the intermediate feature space of a pretrained generative model. 
This transfer function operates as a learnable $1 \times 1$ convolution, enabling precise control over fine-grained detail transfer.
However, generative models are trained at fixed input resolutions and cannot operate directly on high-resolution image setups. 
Thus, to fully inherit their strong generative priors while accommodating large-scale inputs, we adopt a patch-wise strategy, dividing the source image into model-native resolution regions. 
This enables preserving micro-scale details while still benefiting from the priors encoded in the pretrained generative model. 
A novel synchronization mechanism is then applied to ensure global consistency between patches. 
We note that the proposed synchronization method does not require overlapping inference, effectively decreasing the computational cost of the overall method.
We summarize our contributions as follows:
\begin{itemize}[label=$\bullet$]
    \item To the best of our knowledge, our work is the first to propose the high-resolution image editing task, supported by our novel patch-wise inference mechanism.
    \item We develop a feature transfer function and a synchronization strategy across non-overlapping patches to perform high-resolution editing in a test-time optimization manner.
    \item Experimental results demonstrate that the proposed method achieves state-of-the-art editing capacity that enable diverse instruction-based editing.
\end{itemize}

% !TEX root = ./../main.tex

\section{Related work}
\label{sec:related_works}

\paragraph{Text-driven image editing.}
Text-conditional image editing methods inherit the strong generative capabilities of pretrained text-to-image models~\cite{rombach2022high, saharia2022photorealistic, esser2024scaling, flux2024}. 
Building on these priors, approaches using U-Net-based~\cite{ronneberger2015u} diffusion model achieve plausible performance across diverse scenarios via latent optimization, attention manipulation, and instruction-based control~\cite{avrahami2022blended, kim2022diffusionclip, hertz2022prompt, parmar2023zero, tumanyan2023plug, cao2023masactrl, lee2023conditional, couairon2022diffedit, lee2024diffusion, brooks2023instructpix2pix, epstein2023diffusion, lee2025diffusion}.

More recently, generative models with transformer-based architectures~\cite{vaswani2017attention} have further advanced editing capabilities.
For example, Step1X-Edit~\cite{liu2025step1x-edit} presents a general-purpose instruction-based image editing framework by coupling a multi-modal LLM with a DiT-based~\cite{peebles2023scalable} decoder trained on a large-scale taxonomy-driven editing dataset.
ICEdit~\cite{zhang2025context} leverages the in-context generation capability of large DiT models, treating edits as diptych-style prompts and only requires lightweight LoRA-MoE fine-tuning~\cite{hu2022lora}.
KV-Edit~\cite{zhu2025kv} enables training-free editing by reusing cached key–value tokens under a user-provided mask, which is an idea inspired by LLM's KV-cache mechanisms~\cite{xiao2023efficient}, to preserve background pixels while effectively modifying the foreground regions.
However, these methods can edit images only up to 1K resolution, whereas our approach leverages the generative priors of existing methods to enable editing even at 2K and higher resolutions.

\vspace{-4mm}
\paragraph{Diffusion-based super-resolution.}

Real-World Image Super Resolution (Real-ISR) seeks to reconstruct detailed high-resolution images from low-resolution inputs that exhibit diverse and unknown degradations.
With the rapid advancement of diffusion models, diffusion-based approaches have increasingly been adopted to tackle this task.
Prior methods~\cite{wang2024exploiting, wu2024seesr, ai2024dreamclear, wang2024sinsr, yu2024scaling, wu2024one, duan2025dit4sr, cheng2025effective, sun2024pisasr, dong2025tsd, kang2025icm} built on large text-to-image models have recently shown strong performance. 
For instance, TSD-SR~\cite{dong2025tsd} distills a multi-step text-to-image diffusion model into a one-step super-resolution network through target score distillation and distribution-aware sampling.
PiSA-SR~\cite{sun2024pisasr} introduces dual LoRA~\cite{hu2022lora} modules separately dedicated to pixel-level regression and semantic-level enhancement.
On the other hand, DiT-SR~\cite{cheng2025effective} introduces a diffusion transformer~\cite{peebles2023scalable} trained from scratch with a U-shaped multi-scale design and frequency-aware timestep conditioning.
DiT4SR~\cite{duan2025dit4sr} adapts the Stable Diffusion 3~\cite{esser2024scaling} architecture by injecting low-resolution features directly into transformer~\cite{vaswani2017attention} blocks via bidirectional attention and convolution.
While existing SR methods rely on training to generate high-resolution outputs, our method restores fine details through test-time optimization without any model training.

% !TEX root = ./../main.tex

\section{Preliminary}
\label{sec:preliminary}

\paragraph{Diffusion models.} 
Diffusion models~\cite{sohl2015deep, ho2020denoising, song2020denoising, song2020score} demonstrate a remarkable ability to synthesize versatile content in multiple modalities.
Starting from an initial noisy latent $\mathbf{x}_T \sim \mathcal{N}(\mathbf{0}, \mathbf{I})$, they generate a trajectory of latent variables $\{ \mathbf{x}_t^{\mathrm{rev}}\}_{t=T}^0$ through a reverse diffusion process.
DDIM~\cite{song2020denoising} formulates this reverse process as a deterministic procedure.
Specifically, $\mathbf{x}_{t-1}^{\mathrm{rev}}$ is sampled from $\mathbf{x}_t^{\mathrm{rev}}$ as:
\begin{align}
    \mathbf{x}_{t-1}^{\mathrm{rev}}  & = f^{\mathrm{rev}}(\mathbf{x}_t^{\mathrm{rev}}, t)  \nonumber \\ &:=  \sqrt{\alpha_{t-1}} \hat{\mathbf{x}}_0(\mathbf{x}_t^{\mathrm{rev}}, t) + \sqrt{1-\alpha_{t-1}} \epsilon_{\theta}(\mathbf{x}_t^{\mathrm{rev}}, t)
    \label{eq:ddim_reverse}
\end{align}
where $\{\alpha_t \}_{t=0}^T$ is the predefined decreasing variance schedule, $\epsilon_{\theta}(\cdot, \cdot)$ denotes the pretrained noise prediction network, and $\hat{\mathbf{x}}_0(\mathbf{x}_t, t)$ is the estimated Tweedie at timestep $t$, computed with the Tweedie's Formula~\cite{stein1981estimation} given by
\begin{equation}
    \hat{\mathbf{x}}_0(\mathbf{x}_t, t) = \frac{\mathbf{x}_t - \sqrt{1-\alpha_t} \epsilon_{\theta}(\mathbf{x}_t, t)}{\sqrt{\alpha_t}}.
    \label{eq:tweedie}
\end{equation}
Iteratively applying Eq.~\eqref{eq:ddim_reverse} gives a clean data sample $\mathbf{x}_0$.

\paragraph{Image-to-image translation.}
In diffusion-based image-to-image translation~\cite{parmar2023zero, tumanyan2023plug, cao2023masactrl}, deterministic forward and reverse processes~\cite{song2020denoising} are commonly employed to reconstruct a given source image.
Starting from a source image $\mathbf{x}_0$, the forward process introduces noise according to
\begin{align}
    \mathbf{x}_{t+1}^{\mathrm{fwd}}  & = f^{\mathrm{fwd}}(\mathbf{x}_t^{\mathrm{fwd}}, t) \nonumber \\ & := \sqrt{\alpha_{t+1}}  \hat{\mathbf{x}}_0(\mathbf{x}_t^{\mathrm{fwd}}, t) + \sqrt{1-\alpha_{t+1}} \epsilon_{\theta}(\mathbf{x}_t^{\mathrm{fwd}}, t), 
    \label{eq:ddim_inversion}
\end{align}
and the iterative application of Eq.~\eqref{eq:ddim_inversion} for $T$ times yield the noised source latent $\mathbf{x}_T^{\mathrm{fwd}}$. 
The reconstruction of the source image is then obtained by applying the reverse DDIM process (Eq.~\eqref{eq:ddim_reverse}) from $\mathbf{x}_T^{\mathrm{rev}} = \mathbf{x}_T^{\mathrm{fwd}}$, producing $\mathbf{x}_0^{\mathrm{recon}}$.
Under the assumption of negligible discretization error~\cite{song2020score}, this forward–reconstruction procedure is theoretically lossless, guaranteeing $\mathbf{x}_0 = \mathbf{x}_0^{\mathrm{recon}}$.

\paragraph{Null-text inversion.} 
In practice, discretization errors can accumulate during the forward and reverse processes, leading to imperfect reconstructions.
To account for such deviations, Null-text inversion~\cite{mokady2023null} refines the reverse process (Eq.~\eqref{eq:ddim_reverse}) by introducing a timestep-dependent parameter $\phi_t$, which serves as a learnable unconditional embedding.
The parameter $\phi_t$ is optimized to align the reverse trajectory with the forward trajectory by minimizing the discrepancy between their corresponding latents:
\begin{equation}
    \min_{\phi_t} \|\mathbf{x}_{t-1}^{\mathrm{fwd}} - f^{\mathrm{rev}}(\mathbf{\tilde{x}}_{t}^{\mathrm{rev}}, t, \phi_{t}) \|^2_2,
\end{equation}
where $\tilde{\mathbf{x}}_{t}^{\mathrm{rev}}$ denotes the latent obtained from the optimized reverse process. 
This strategy enables an accurate reconstruction of the source image.

% !TEX root = ./../main.tex

\section{{\methodname}}
\label{sec:method}

\begin{figure*}[t!]
	\centering
	\includegraphics[width=1.0\linewidth]{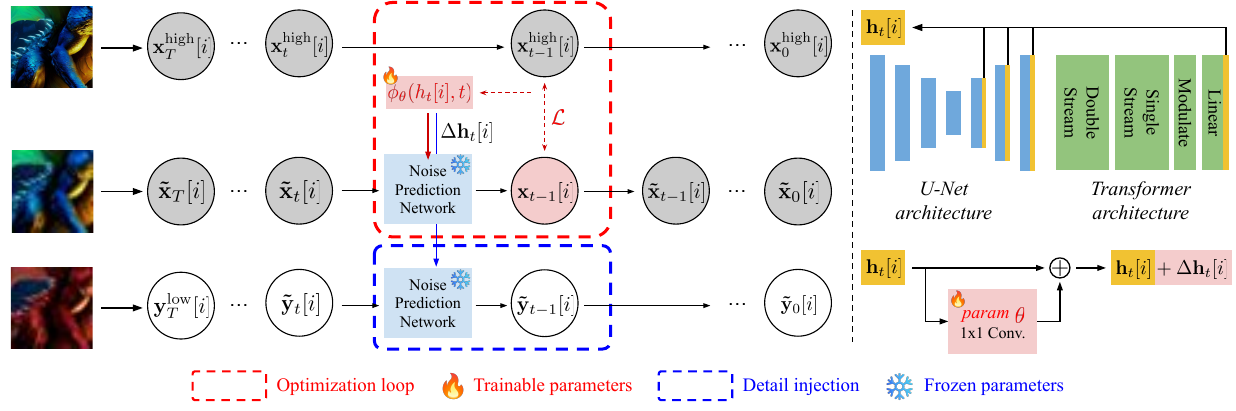}
    \vspace{-7mm}
	\caption{
    Overview of the proposed method. 
    (Left) We first optimize the transfer function $\phi_{\theta}(\mathbf{h}_i[t], t)$ to capture fine-grained details encoded in the high-resolution source trajectory $\{\srch_t[i] \}_{t=0}^{T}$. 
    We then apply the optimized transfer function during the reverse process of $\{ \mathbf{\tilde{y}}_t [i]\}_{t=0}^{T}$, yielding a detail-enhanced latent $\mathbf{\tilde{y}}_0[i] = \mathbf{y}_{0}^{\mathrm{high}}[i].$
    (Right) We illustrate how the transfer function modulates the intermediate feature within the pretrained model.
    }  
	\vspace{-5mm}
\label{fig:method}
\end{figure*}

\subsection{Problem formulation}

We consider three conditioning images: 
(1) the high-resolution source $\imghs \in \mathbb{R}^{H_h \times W_h \times 3}$, 
(2) its downsampled version $\imgls \in \mathbb{R}^{H_l \times W_l \times 3}$, 
(3) and a low-resolution reference $\imglr \in \mathbb{R}^{H_l \times W_l \times 3}$ that represents the desired target image.
The image $\imgls$ is obtained by downsampling $\imghs$, and the reference $\imglr$ is produced by applying a standard low-resolution image editing method (\textit{e.g.} Nano Banana~\cite{comanici2025gemini}) to $\imgls$.
The objective is to generate a high-resolution edited output $\imghr$ that reflects the overall semantics of $\imglr$ while maintaining the fine-grained details present in $\imghs$.

\subsection{Overview}
\label{subsec:method_overview}
The proposed method generates high-resolution output $\imghr$ in a patch-wise manner, for precise control over the fine-grained details present in $\imghs$.
We first divide all conditioning images into $N \times M$ patches (Sec.~\ref{subsec:method_patch}).
For each patch, we transfer the fine-grained details residing in $\imghs$ by manipulating the intermediate features of a pretrained generative model (Sec.~\ref{subsec:method_hf}), producing $N \times M$ enhanced target patches.
Then, they are synchronized and fused into a coherent high-resolution target image (Sec.~\ref{subsec:method_sync}), ensuring smoothness across patch boundaries.
Figure~\ref{fig:method} provides a summary of our method.

\subsection{Patch-wise generation trajectory extraction}
\label{subsec:method_patch}

Given three conditioning images, we first resize the low-resolution source $\imgls$ and reference image $\imglr$ to match the resolution of the high-resolution source $\imghs$.
We then divide each image into non-overlapping $N \times M$ patches of size $\mathbb{R}^{H_d \times W_d \times 3}$, where $H_d \times W_d$ corresponds to the input resolution of the pretrained generative model~\cite{rombach2022high, flux2024}.
Thus, $N = H_h / H_d$ and $M = W_h / W_d$, resulting in a total of $N \times M$ patches.
We denote the $i^\text{th}$ patch of $\imghs$ as $\imghs[i]$, and we define $\imgls[i]$ and $\imglr[i]$ likewise.

Each patch is then encoded using the pretrained VAE~\cite{kingma2013auto} encoder.
Here, we define the resulting latent representations of the high-resolution source, low-resolution source, and low-resolution reference patches as
$\srch_0[i]$, $\srcl_0[i]$, and $\refl_0[i]$, respectively, where $1 \leq i \leq NM$.
We subsequently apply the forward process, optionally enhanced with Null-text inversion~\cite{mokady2023null}, to obtain their generation trajectories, which are given by
\begin{equation}
    \{\srch_t[i] \}_{t=0}^{T}, \{\srcl_t[i] \}_{t=0}^{T}, \{\refl_t [i]\}_{t=0}^{T}.
\end{equation}

\subsection{Detail enhancement module}
\label{subsec:method_hf}

To inject fine-grained details from a low-resolution image to a target high-resolution image, we introduce a transfer function that aligns the generation trajectories of the low-resolution into high-resolution domains.
Specifically, starting from $\mathbf{\tilde{x}}_T[i]=\srcl_T[i]$, we aim to guide its generation trajectory $\{\mathbf{\tilde{x}}_t [i]  \}_{t=0}^{T}$ to follow that of the high-resolution source latent $\{\srch_t [i] \}_{t=0}^{T}$.

To accomplish this, we define a timestep-dependent transfer function $\phi(i, t)$ that adjusts the intermediate feature representation $\mathbf{h}_t[i]$ of the pretrained generative model~\cite{rombach2022high, flux2024}.
Here, $\mathbf{h}_t[i]$ denotes the output of the ResNet block~\cite{he2016deep} along the upward path of the U-Net architecture~\cite{ronneberger2015u}, or the output of the final linear layer following the single-stream block in the transformer-based architecture~\cite{flux2024}.
This is illustrated as the yellow sub-block on the right side of Figure~\ref{fig:method}.
We then introduce a feature modification term $\Delta \mathbf{h}_t[i] = \phi(i, t)$, which is applied to manipulate the intermediate feature of the pretrained model at each timestep $t$.
The transfer function is optimized to align the low-resolution trajectory with the high-resolution trajectory by minimizing the following training objective:
\begin{equation}
\mathcal{L}:=  ||\srch_{t-1}[i] - f^{\mathrm{rev}}(\mathbf{\tilde{x}}_t[i], t ~; \Delta \mathbf{h}_t[i]) ||^2_2,
\label{eq:objective_delta_h}
\end{equation}
where $f^{\mathrm{rev}}(\cdot, \cdot ~; \Delta \mathbf{h}_t[i])$ denotes the reverse process with the manipulated intermediate feature of $\mathbf{h}_t[i] + \Delta \mathbf{h}_t[i]$.
For DDIM~\cite{song2020denoising}, it is defined as follows:
\begin{align}
    f^{\mathrm{rev}}(\mathbf{x}_t^{\mathrm{rev}}, t ~; \Delta \mathbf{h}_t[i])   &:=  \sqrt{\alpha_{t-1}} \hat{\mathbf{x}}_0(\mathbf{x}_t^{\mathrm{rev}}, t ~; \Delta \mathbf{h}_t[i])
    \label{eq:ddim_reverse_with_h} 
    \\ & + \sqrt{1-\alpha_{t-1}} \epsilon_{\theta}(\mathbf{x}_t^{\mathrm{rev}}, t ~; \Delta \mathbf{h}_t[i]). \nonumber 
\end{align}
Note that  $f^{\mathrm{rev}}(\mathbf{x}_t^{\mathrm{rev}}, t )$ in Eq.~\eqref{eq:ddim_reverse} is equal to $ f^{\mathrm{rev}}(\mathbf{x}_t^{\mathrm{rev}}, t ~; \mathbf{0})$.

A straightforward parameterization of the timestep-dependent transfer function is to use a learnable constant vector $\phi(i, t) = \mathbf{c}_t[i]$, resulting in the adjusted intermediate feature $\mathbf{h}_t[i] + \mathbf{c}_t[i]$.
However, we observe that this parameterization fails to produce reliable transformations, especially when the source and reference differ in semantics, \textit{e.g.}, cat-to-dog.
We hypothesize that this limitation arises because a single global constant vector is insufficient to model varying degrees of change throughout the edited image, and thus cannot adapt to diverse spatial structures.

\begin{figure*}[t!]
	\centering
    \includegraphics[width=\linewidth]{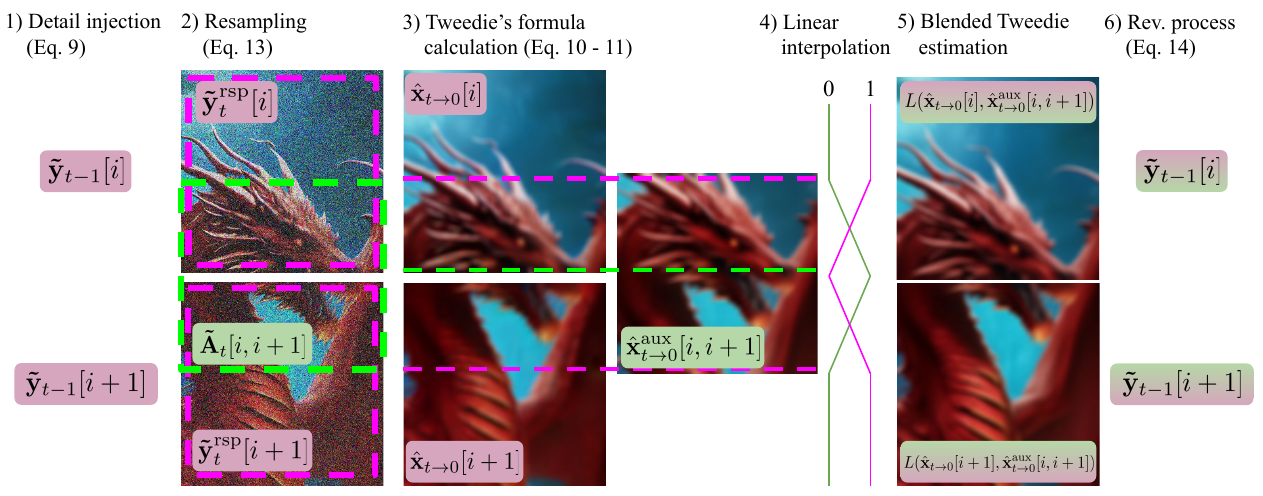}
    \vspace{-4mm}
	\caption{
    Overview of the synchronization strategy.
    Starting from the detail enhancement process (Eq.~\eqref{eq:ddim_reverse_detail_inject}), we perform resampling (Eq.~\eqref{eq:resample_inversion}) and compute the blended Tweedie estimate. 
    This estimate is then used to synchronize adjacent patches during the reverse process (Eq.~\eqref{eq:resample_reverse}). 
    }  
	\vspace{-4mm}
\label{fig:method_sync}
\end{figure*}

To overcome this issue, we instead formulate the transfer function as an operation adaptive to $\mathbf{h}_t[i]$, rather than as a global constant vector.
That is, we design the feature modification term as
\begin{equation}
    \Delta \mathbf{h}_t[i] :=\phi_{\theta}(\mathbf{h}_t[i], t),
    \label{eq:delta_h_def}
\end{equation}
where $\theta$ is the trainable parameter that defines the operation for each timestep $t$.
Empirically, we define the transfer function using a $1 \times 1$ convolution applied to $\mathbf{h}_t[i]$, \textit{i.e.} $\mathrm{conv}_{1 \times 1}(\mathbf{h}_t[i])$. 
This lightweight operation allows feature mixing across channels while preserving spatial layout, which turns out to be effective for transferring fine-grained details in a spatially adaptive manner.
A detailed discussion of the design of transfer function is explained in Appendix.

After optimizing the transfer function, we incorporate it during the reverse process to generate the target image, starting from $ \mathbf{\tilde{y}}_T[i] = \refl_T[i]$ and resulting in a trajectory of $\{ \mathbf{\tilde{y}}_t [i]\}_{t=0}^{T}$, where $\mathbf{\tilde{y}}_{t-1} [i]$ is sampled via
\begin{equation}
    \mathbf{\tilde{y}}_{t-1} [i] = f^{\mathrm{rev}}(\mathbf{\tilde{y}}_t[i], t ~; \Delta \mathbf{h}_t[i]).
    \label{eq:ddim_reverse_detail_inject}
\end{equation}
Then, we finally take $\tgth_0[i] = \mathbf{\tilde{y}}_0[i]$.
This sampling strategy is effective since the objective in Eq.~\eqref{eq:objective_delta_h} encourages the transfer function to act as a feature-level detail injection operator, guiding the low-resolution source trajectory toward the high-resolution source trajectory while preserving the fine-grained details.

Optimizing and injecting $\Delta \mathbf{h}_t[i]$ over all timesteps facilitates fine-detail transfer, but may gradually distort the content of the source image and yield higher computation.
In contrast, limiting optimization to early timesteps preserves the source image's content but may under-transfer fine details.
To handle this trade-off, we introduce a hyperparameter $\tau$ that denotes the initial timestep at which our transfer function optimization is applied; 
that is, $\Delta \mathbf{h}_t[i] = \mathbf{0}, \ ^\forall t > \tau$.
Thus, it serves as a flexible controller that balances the level of detail transfer relative to content preservation, further allowing one to choose the trade-off depending on the editing strength and desired fidelity.

\subsection{Synchronization between patches}
\label{subsec:method_sync}

While the transfer function $\Delta \mathbf{h}_t[i]$ injects fine-grained details into each patch, applying it independently to each patch may lead to inconsistent textures and artifacts at patch boundaries.
To ensure global coherence across patches, we introduce a synchronization mechanism that combines blended Tweedie updates (Sec.~\ref{subsubsec:tweedie}) with a resampling strategy (Sec.~\ref{subsubsec:resample}).
For clarity, we describe the case where adjacent patches are vertically aligned within the diffusion process, namely $\mathbf{\tilde{y}}_t[i]$ and $\mathbf{\tilde{y}}_t[i+1]$.
The same procedure applies equivalently when they are placed along the horizontal directions.

\subsubsection{Blended-Tweedie-based updates}
\label{subsubsec:tweedie}
To synchronize adjacent patches, we introduce an auxiliary latent $\mathbf{\tilde{A}}_t [i, i+1]$, constructed by spatially blending the bottom half of $\mathbf{\tilde{y}}_t[i]$ and the upper half of $\mathbf{\tilde{y}}_t[i+1]$ at timestep $t$.
Since each patch is denoised independently, their latent trajectories may diverge, often producing visible discontinuities along their boundaries.
However, $\mathbf{\tilde{A}}_t[i, i+1]$ contains the boundary region, and therefore its Tweedie estimate may naturally captures a smoother transition between the two patches.
By blending the Tweedie estimate of an auxiliary latent with those of the original patches, the boundary between neighboring patches becomes more coherent, reducing artifacts and enhancing spatial continuity.

For spatial blending, we first apply Eq.~\eqref{eq:tweedie} to obtain the original patches' Tweedie estimates:
\begin{align}
    \hat{\mathbf{x}}_{t \rightarrow 0}^{\mathrm{aux}}[i, i+1] &= \hat{\mathbf{x}}_0 (\mathbf{\tilde{A}}_t[i, i+1], t) \\
    \hat{\mathbf{x}}_{t \rightarrow 0}[i] &= \hat{\mathbf{x}}_0 (\mathbf{\tilde{y}}_t[i], t).
\end{align}
Then we define $L(\hat{\mathbf{x}}_{t \rightarrow 0}[i], \hat{\mathbf{x}}_{t \rightarrow 0}^{\mathrm{aux}}[i, i+1])$, a blended Tweedie estimate that is used to swap the $\hat{\mathbf{x}}_0$ term in Eq.~\eqref{eq:ddim_reverse} during the reverse process of $\mathbf{\tilde{y}}_t[i]$.
Let $W_p$ and $H_p$ denote the width and height of a single patch, respectively.
We linearly interpolate the bottom half of $\mathbf{\tilde{y}}_t[i]$ and the corresponding region of the auxiliary latent $\mathbf{\tilde{A}}_t[i, i+1]$, \textit{i.e.} the top half of $\mathbf{\tilde{A}}_t[i, i+1]$, whose overlap has spatial size $H_p / 2 \times  W_p$.
To achieve this, we define a vertical interpolation weight as:
\begin{equation}
    \mathbf{M}(v, t) = \frac{2v} {H_p} \cdot \left(1 - \frac{t}{\tau} \right) \quad 0 \leq v \leq H_p / 2,
\end{equation}
which increases linearly from the boundary toward the center of the overlap, so that the contribution of the auxiliary latent grows smoothly across the overlapping region.
We then substitute the bottom half of the Tweedie estimate for $\mathbf{\tilde{y}}_t[i]$ with the linearly interpolated one, while keeping its top half unchanged.
This yields the blended Tweedie estimate with respect to $\hat{\mathbf{x}}_{t \rightarrow 0}[i]$ and $\hat{\mathbf{x}}_{t \rightarrow 0}^{\mathrm{aux}}[i, i+1]$.

For the adjacent patch $\mathbf{\tilde{y}}_t[i+1]$, we apply the vertically mirrored blend operation. 
Specifically, we compute $L(\hat{\mathbf{x}}_{t \rightarrow 0}[i+1], \hat{\mathbf{x}}_{t \rightarrow 0}^{\mathrm{aux}}[i, i+1])$ with the weight of $1-\mathbf{M}$ (vertically reversed ramp).
This blending ensures spatial smoothness between adjacent patches.
Detailed formulas indicating these procedures are described in the Appendix.

\begin{table*}[t]
    \centering
    \caption{Quantitative evaluation under 1K- and 2K-editing scenarios using the pretrained Stable Diffusion~\cite{rombach2022high}. 
    We compare the proposed method with diffusion-based super-resolution methods~\cite{cheng2025effective, duan2025dit4sr, sun2024pisasr, dong2025tsd}.
    Our method shows superior performance compared to the baselines.
    }
    \vspace{-2mm}
    \setlength{\tabcolsep}{4.5pt}
    \scalebox{0.85}{
        \begin{tabular}{lcccccccccc}
            \toprule
            \multirow{2}{*}{Method} 
            & \multicolumn{5}{c}{1K-editing} &  \multicolumn{5}{c}{2K-editing} \\
            \cmidrule(lr){2-6} \cmidrule{7-11} 
            & HaarPSI $\uparrow$  & M-MSE $\downarrow$  & M-SSIM  $\uparrow$
            & M-PSNR $\uparrow$ & LPIPS  $\downarrow$ & HaarPSI $\uparrow$   & M-MSE $\downarrow$   & M-SSIM $\uparrow$ & M-PSNR $\uparrow$ & LPIPS  $\downarrow$   \\
            \midrule
            DiT-SR~\cite{cheng2025effective} & \underline{0.335} 	&\underline{0.058}&	\underline{0.695}	&\underline{21.528} &	0.477 & \underline{0.316} &0.057	&0.754	&\underline{21.380} & 0.507 \\
            DiT4SR~\cite{duan2025dit4sr}   & 0.324&0.060	&0.625&	20.740& 0.509  & 0.305 &0.058&	0.684&	20.701 & 0.534 \\
            PiSA-SR~\cite{sun2024pisasr} & 0.328	&	\underline{0.058}&	0.668	&21.273  & \underline{0.465} &0.312 & \underline{0.056}	&\underline{0.755}	&21.320  & \textbf{0.472} \\
            TSD-SR~\cite{dong2025tsd} & 0.329	&0.061&	0.649	&20.766  &0.489	& 0.312 &0.059&	0.715&	20.796  &0.514\\
            \textbf{\methodname ~(Ours)}  & \textbf{0.342} & \textbf{0.054} & \textbf{0.739} & \textbf{22.132} & \textbf{0.460} & \textbf{0.331} & \textbf{0.053} & \textbf{0.806} & \textbf{21.955} & \underline{0.496} \\
            \bottomrule
        \end{tabular}
        }
    \label{tab:quant}
    \vspace{-2mm}
\end{table*}

\vspace{0mm}
\subsubsection{Resampling strategy}
\label{subsubsec:resample}
A challenge in the above formulation is that $\Delta \mathbf{h}_t$ remains undefined for $\mathbf{\tilde{A}}_t[i, i+1]$, since adjacent patches $\mathbf{\tilde{y}}_t[i]$ and $\mathbf{\tilde{y}}_t[i+1]$ contain independently injected detail terms, $\Delta \mathbf{h}_t[i]$ and $\Delta \mathbf{h}_t[i+1]$.
In principle, $\Delta \mathbf{h}_t$ for $\mathbf{\tilde{A}}_t[i, i+1]$ can be computed explicitly; however, this entails a costly additional optimization.
Instead, we propose an alternative approach which we refer to as the \emph{resampling strategy}, where synchronization is decoupled from detail injection. 
The key idea is to reconstruct latents that preserve the fine-grained details already injected by the transfer function, but without dependency on any explicit $\Delta \mathbf{h}_t$ term.
This allows synchronization to proceed without defining a separate $\Delta \mathbf{h}_t$ for $\mathbf{\tilde{A}}_t[i, i+1]$.

Given the detail-injected latents $\mathbf{\tilde{y}}_{t-1} [i]$ and $\mathbf{\tilde{y}}_{t-1} [i+1]$ obtained via Eq.~\eqref{eq:ddim_reverse_detail_inject}, we apply a forward process \emph{without} re-injecting the transfer function to get the resampled latents $\mathbf{\tilde{y}}_{t}^{\mathrm{rsp}} [i]$ and $\mathbf{\tilde{y}}_{t}^{\mathrm{rsp}} [i+1]$: 
\begin{align}
    \mathbf{\tilde{y}}_{t}^{\mathrm{rsp}} [i] & = f^{\mathrm{fwd}} (\mathbf{\tilde{y}}_{t-1} [i], t-1), \nonumber  \\ \mathbf{\tilde{y}}_{t}^{\mathrm{rsp}} [i+1] & = f^{\mathrm{fwd}} (\mathbf{\tilde{y}}_{t-1} [i+1], t-1).
    \label{eq:resample_inversion}
\end{align}
This resampling process produces detail-preserving latents $\mathbf{\tilde{y}}_{t}^{\mathrm{rsp}}$ that eliminates the need to define $\Delta\mathbf{h}_t$ for the auxiliary latent $\mathbf{\tilde{A}}_t[i,i+1]$.
Using these resampled latents, we compute the blended Tweedie estimate using the synchronization procedure described in Sec.~\ref{subsubsec:tweedie}.
Finally, the actual reverse process at timestep $t$ is performed via Eq.~\eqref{eq:resample_reverse}:
\begin{align}
    \mathbf{\tilde{y}}_{t-1} [i] & = \sqrt{\alpha_{t-1}}  L(\hat{\mathbf{x}}_{t \rightarrow 0}[i], \hat{\mathbf{x}}_{t \rightarrow 0}^{\mathrm{aux}}[i, i+1]) \nonumber \\ &+ \sqrt{1-\alpha_{t-1}} \epsilon_{\theta}(\mathbf{\tilde{y}}_{t}^{\mathrm{rsp}} [i], t). 
    \label{eq:resample_reverse}
\end{align}
The proposed mechanism harmonizes patch-wise diffusion trajectories, yielding globally consistent synthesis with reduced boundary artifacts and preserved fine-grained details.

% !TEX root = ./../main.tex

\section{Experiments}
\label{sec:exp}

\subsection{Implementation details}
\label{subsec:exp_imple}

We implement our method based on the official codebase of Null-text inversion~\cite{mokady2023null} using PyTorch~\cite{paszke2019pytorch}.
We primarily evaluate our method on Stable Diffusion~\cite{rombach2022high} v2.1-base and further demonstrate its applicability to FLUX.1-dev~\cite{flux2024}, which adopts a transformer-based architecture~\cite{vaswani2017attention}.
For both cases, we use a total timestep of $T=50$ and set hyperparameter $\tau=15$.
During the forward and reverse process, we use an empty prompt (``''), since a global text prompt describing $\imghs$ and $\imglr$ does not fully cover the semantics of each patch. 
We adapt Null-text inversion for Stable Diffusion to achieve accurate reconstruction.
After obtaining $N \times M$ latents $\{ \tgth_0[i] \}_{i=1}^{NM}$, we spatially merge them and decode the resulting latent using the pretrained VAE~\cite{kingma2013auto} decoder to get $\imghr$.
Additional implementation details are provided in Appendix.

\subsection{Data acquisition}
\label{subsec:exp_data}

\paragraph{Editing scenarios.}
To comprehensively evaluate high-resolution image editing, we consider two settings that reflect practical usage scenarios.
We first examine the \emph{1K-editing} scenario, where the reference and target resolutions are set to $512^2$ and 1K, respectively.
While existing editing methods generally support editing up to this scale, we include this setting not as a challenging benchmark, but to highlight that our method transfers the fine-grained details faithfully.
Next, we evaluate \emph{2K-editing}, which corresponds to a more common real-world setup for editing beyond the operating resolution of pretrained generative models.

\vspace{-10pt}
\paragraph{Asset generation.}
We construct high-quality assets tailored for our task.
We first generate a total of 100 source images at 4K resolution using FreeScale~\cite{qiu2025freescale}, then downsample them to 2K, 1K, and $512^2$ resolution to obtain multi-resolution source images.
Source prompts and editing instructions are produced with ChatGPT, where the instructions span four editing configurations: 
(1) two types of object replacement, (2) style transfer, and (3) background modification.
We then apply Nano Banana~\cite{comanici2025gemini} to the $512^2$ and 1K versions of each source image to obtain the corresponding edited images.
Here, the $512^2$-sized images serve as low-resolution references for 1K-editing, whereas the 1K resolution images serve as references for 2K-editing.

% !TEX root = ./../main.tex

\begin{figure*}[!t]
	\centering
	\setlength{\tabcolsep}{0mm}
	\renewcommand{\arraystretch}{0.7}
	\hspace{-3mm}
	\scalebox{0.95}{
		\begin{tabular}{c}
			\begin{minipage}{\linewidth}
				\centering
				\begin{tabular}{c@{\extracolsep{1mm}}c@{\extracolsep{1mm}}c@{\extracolsep{1mm}}c@{\extracolsep{1mm}}c@{\extracolsep{1mm}}c@{\extracolsep{1mm}}c@{\extracolsep{1mm}}c@{\extracolsep{1mm}}c@{\extracolsep{1mm}}c}
					\multicolumn{1}{c}{Source (1K)} &
					\multicolumn{1}{c}{Reference ($512^2$)} &
					\multicolumn{2}{c}{Zoomed} &
					\multicolumn{1}{c}{Ours} &
					\multicolumn{1}{c}{PiSA-SR} &
					\multicolumn{1}{c}{DiT-SR} &
					\multicolumn{1}{c}{DiT4SR} &
					\multicolumn{1}{c}{TSD-SR} \\[1mm]
					\includegraphics[height=2.47cm]{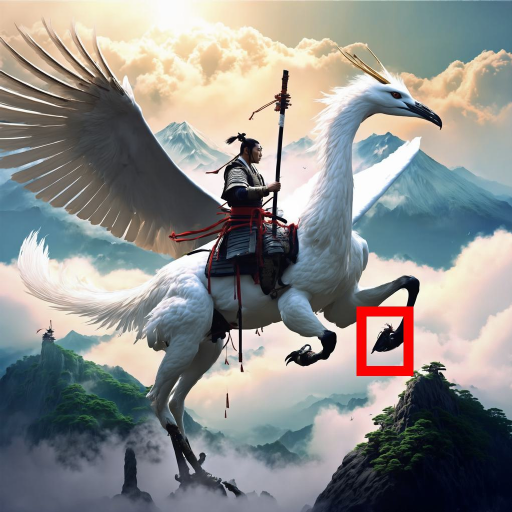} &
					\includegraphics[height=2.47cm]{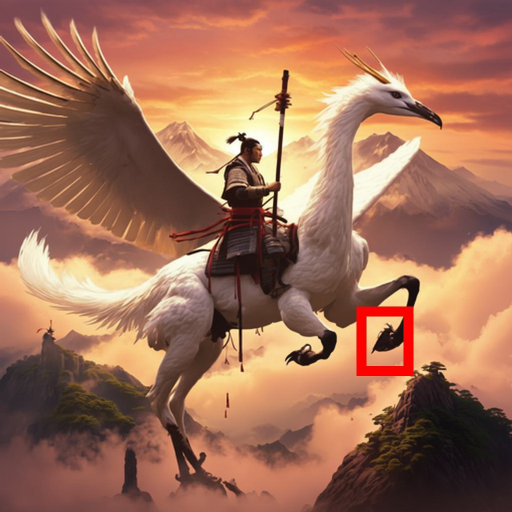} &
					\includegraphics[height=2.47cm]{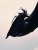} &
					\vdashline &
					\includegraphics[height=2.47cm]{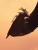} &
					\includegraphics[height=2.47cm]{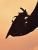} &
					\includegraphics[height=2.47cm]{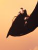} &
					\includegraphics[height=2.47cm]{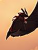} &
					\includegraphics[height=2.47cm]{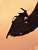} \\
					\multicolumn{9}{c}{\textit{``Change the background to a sunset sky.''}}
				\end{tabular}
			\end{minipage} \\[3mm]
			
			\begin{minipage}{\linewidth}
				\centering
				\begin{tabular}{c@{\extracolsep{1mm}}c@{\extracolsep{1mm}}c@{\extracolsep{1mm}}c@{\extracolsep{1mm}}c@{\extracolsep{1mm}}c@{\extracolsep{1mm}}c@{\extracolsep{1mm}}c@{\extracolsep{1mm}}c}
					\includegraphics[height=2.47cm]{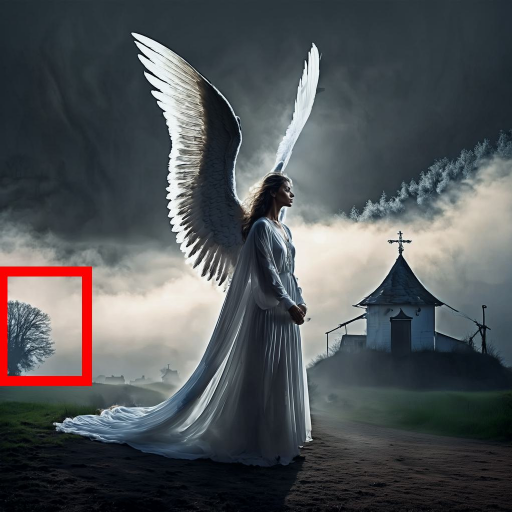} &
					\includegraphics[height=2.47cm]{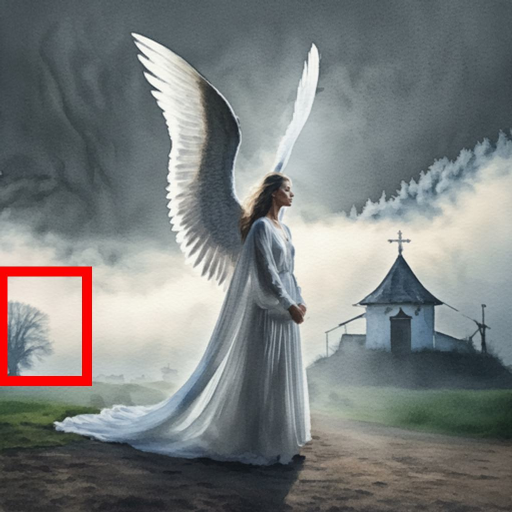} &
					\includegraphics[height=2.47cm]{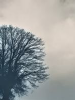} &
					\vdashline &
					\includegraphics[height=2.47cm]{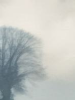} &
					\includegraphics[height=2.47cm]{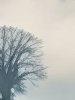} &
					\includegraphics[height=2.47cm]{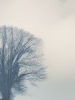} &
					\includegraphics[height=2.47cm]{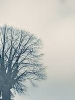} &
					\includegraphics[height=2.47cm]{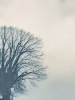} \\
					\multicolumn{9}{c}{\textit{``Apply a soft watercolor style.''}}
				\end{tabular}
			\end{minipage} \\[3mm]
			
			\begin{minipage}{\linewidth}
				\centering
				\begin{tabular}{c@{\extracolsep{1mm}}c@{\extracolsep{1mm}}c@{\extracolsep{1mm}}c@{\extracolsep{1mm}}c@{\extracolsep{1mm}}c@{\extracolsep{1mm}}c@{\extracolsep{1mm}}c@{\extracolsep{1mm}}c}
					\includegraphics[height=2.47cm]{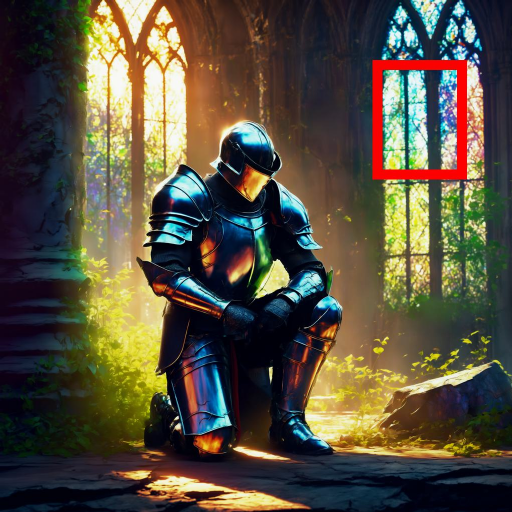} &
					\includegraphics[height=2.47cm]{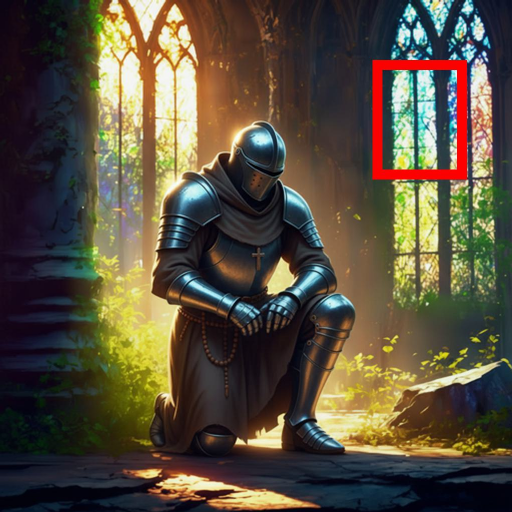} &
					\includegraphics[height=2.47cm]{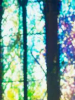} &
					\vdashline &
					\includegraphics[height=2.47cm]{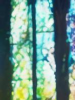} &
					\includegraphics[height=2.47cm]{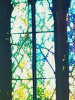} &
					\includegraphics[height=2.47cm]{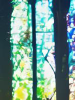} &
					\includegraphics[height=2.47cm]{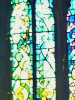} &
					\includegraphics[height=2.47cm]{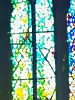}  \\
					\multicolumn{9}{c}{\textit{``Replace the man with a monk wearing armor.''}}
				\end{tabular}
			\end{minipage} \\[3mm]
			
			\\
			
			\begin{minipage}{\linewidth}
				\centering
				\begin{tabular}{c@{\extracolsep{1mm}}c@{\extracolsep{1mm}}c@{\extracolsep{1mm}}c@{\extracolsep{1mm}}c@{\extracolsep{1mm}}c@{\extracolsep{1mm}}c@{\extracolsep{1mm}}c@{\extracolsep{1mm}}c}
					\multicolumn{1}{c}{Source (2K)} &
					\multicolumn{1}{c}{Reference (1K)} &
					\multicolumn{2}{c}{Zoomed} &
					\multicolumn{1}{c}{Ours} &
					\multicolumn{1}{c}{PiSA-SR} &
					\multicolumn{1}{c}{DiT-SR} &
					\multicolumn{1}{c}{DiT4SR} &
					\multicolumn{1}{c}{TSD-SR}  \\[1mm]
					\includegraphics[height=2.47cm]{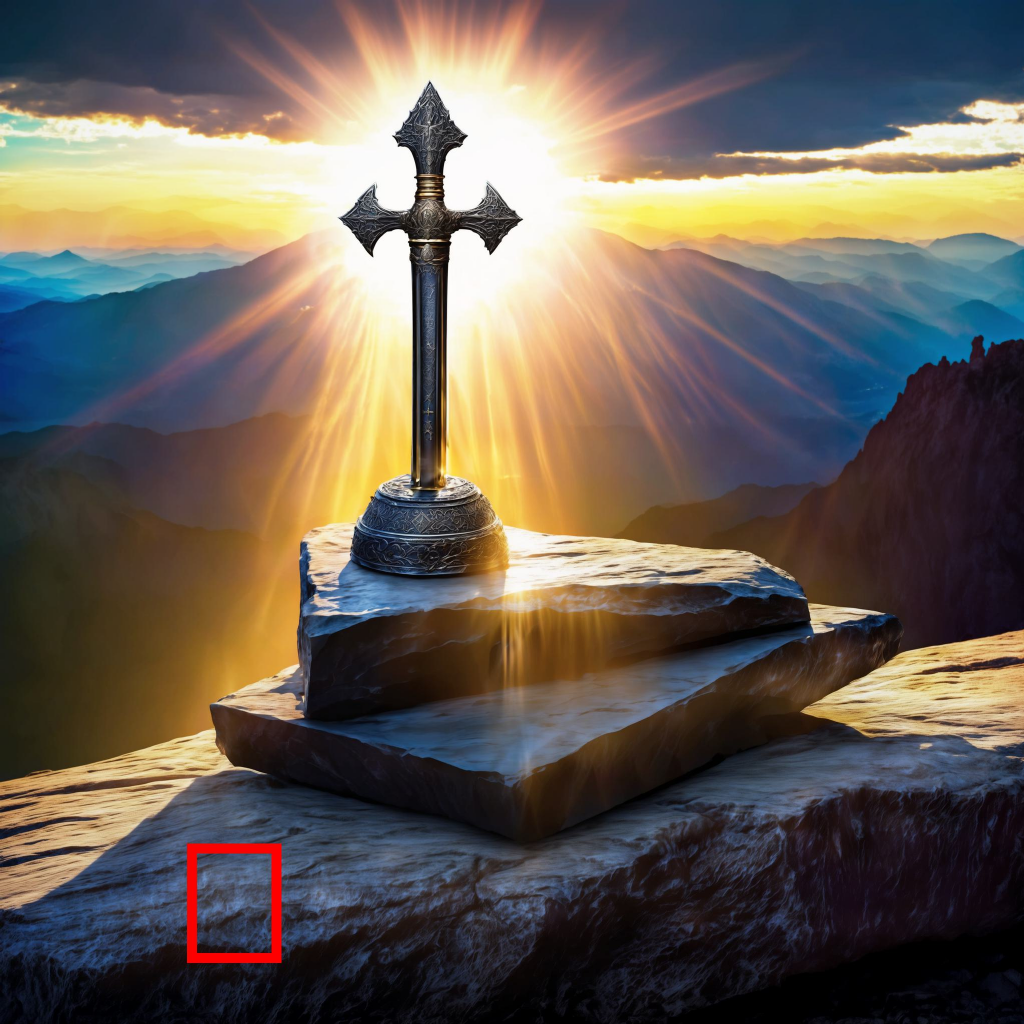} &
					\includegraphics[height=2.47cm]{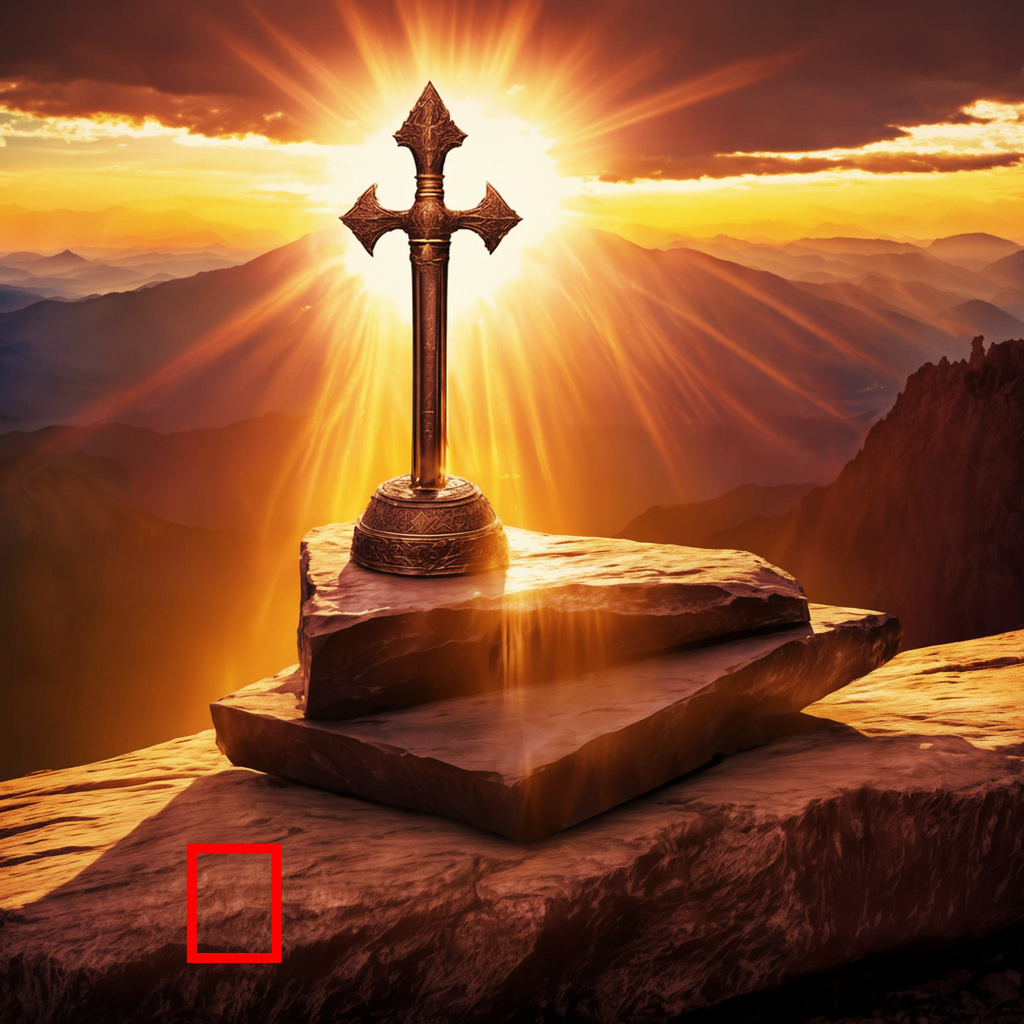} &
					\includegraphics[height=2.47cm]{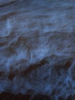}  &
					\vdashline &
					\includegraphics[height=2.47cm]{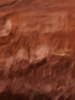} &
					\includegraphics[height=2.47cm]{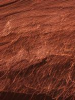} &
					\includegraphics[height=2.47cm]{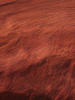} &
					\includegraphics[height=2.47cm]{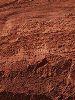} &
					\includegraphics[height=2.47cm]{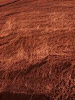} \\
					\multicolumn{9}{c}{\textit{``Apply a warm photo style.''}}
				\end{tabular}
			\end{minipage} \\[3mm]

			\begin{minipage}{\linewidth}
				\centering
				\begin{tabular}{c@{\extracolsep{1mm}}c@{\extracolsep{1mm}}c@{\extracolsep{1mm}}c@{\extracolsep{1mm}}c@{\extracolsep{1mm}}c@{\extracolsep{1mm}}c@{\extracolsep{1mm}}c@{\extracolsep{1mm}}c}
					\includegraphics[height=2.47cm]{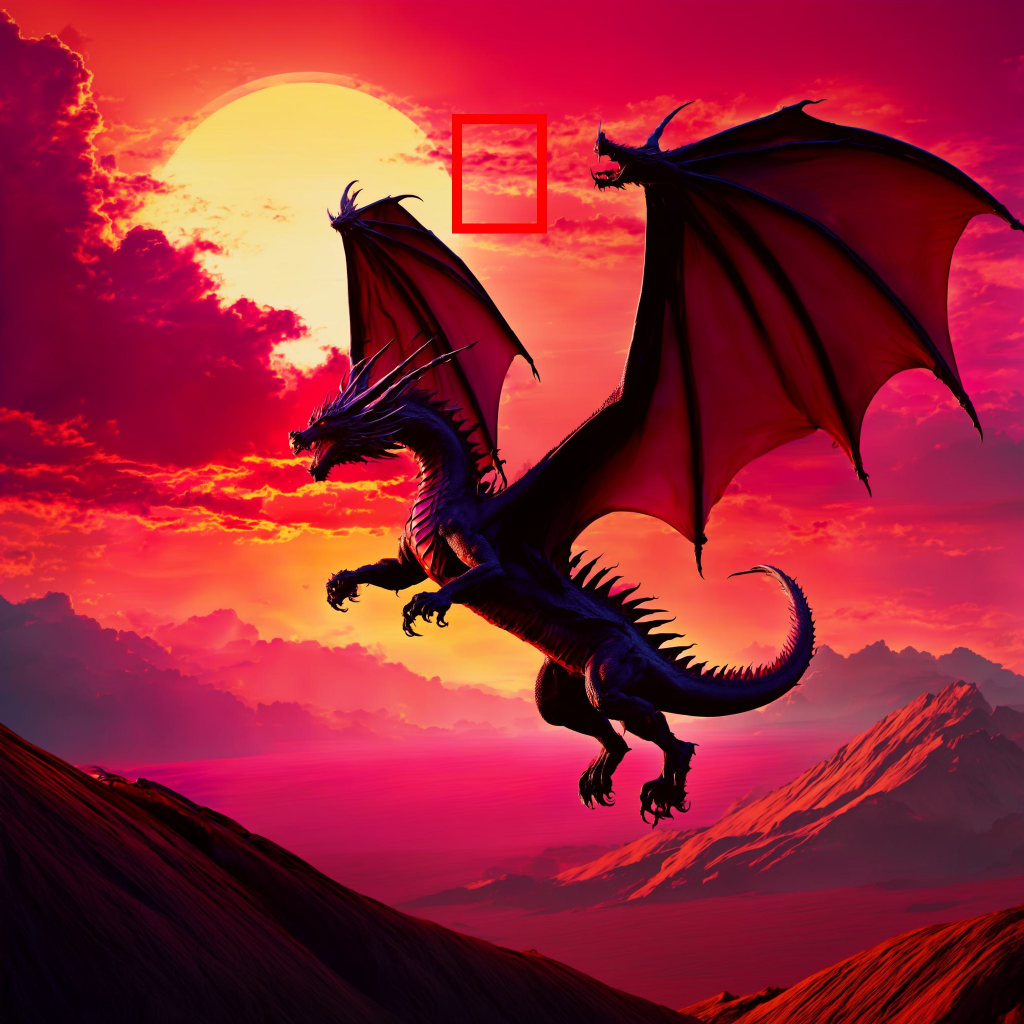} &
					\includegraphics[height=2.47cm]{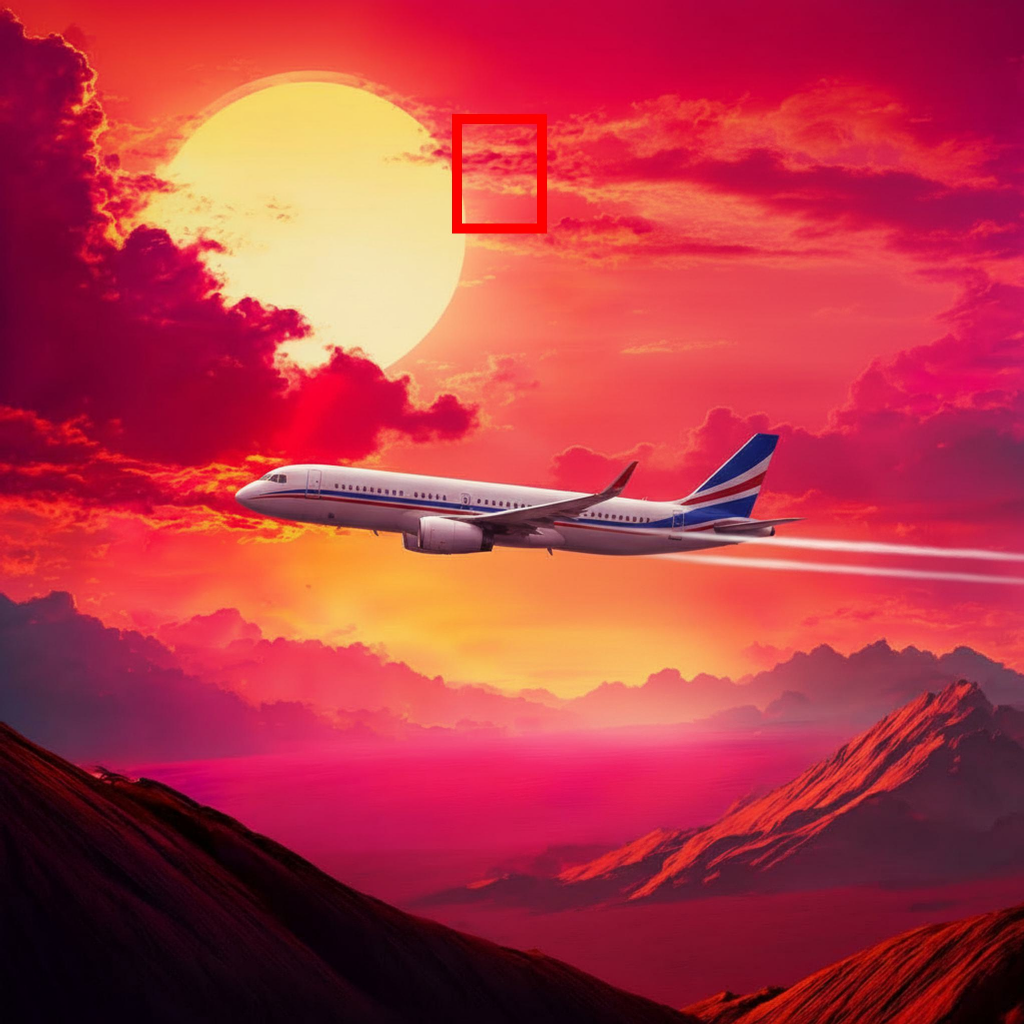} &
					\includegraphics[height=2.47cm]{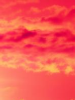}  &
					\vdashline &
					\includegraphics[height=2.47cm]{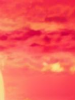} &
					\includegraphics[height=2.47cm]{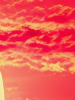} &
					\includegraphics[height=2.47cm]{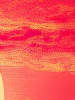} &
					\includegraphics[height=2.47cm]{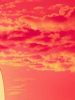} &
					\includegraphics[height=2.47cm]{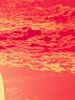} \\
					\multicolumn{9}{c}{\textit{``Replace the dragon with a plane flying through the clouds.''}}
				\end{tabular}
			\end{minipage} \\[3mm]
			
			\begin{minipage}{\linewidth}
				\centering
				\begin{tabular}{c@{\extracolsep{1mm}}c@{\extracolsep{1mm}}c@{\extracolsep{1mm}}c@{\extracolsep{1mm}}c@{\extracolsep{1mm}}c@{\extracolsep{1mm}}c@{\extracolsep{1mm}}c@{\extracolsep{1mm}}c}
					\includegraphics[height=2.47cm]{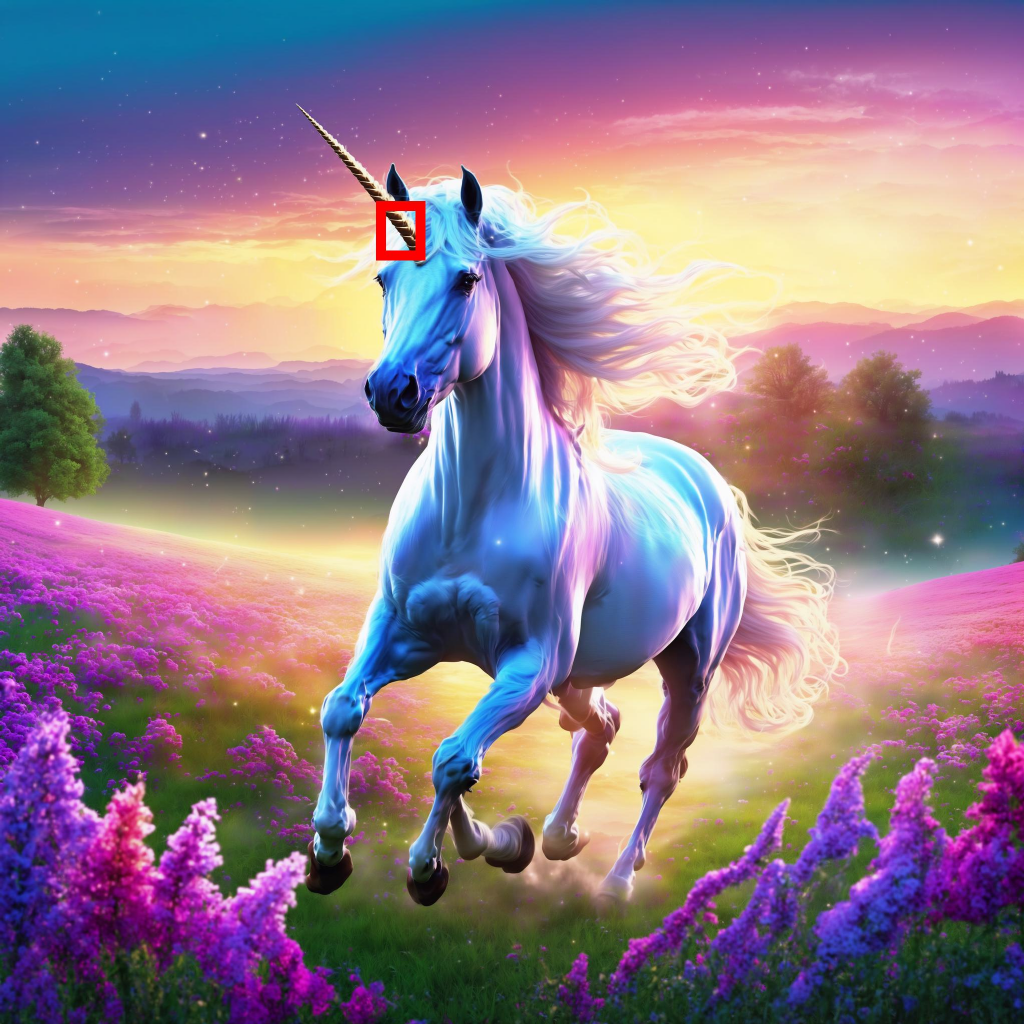} &
					\includegraphics[height=2.47cm]{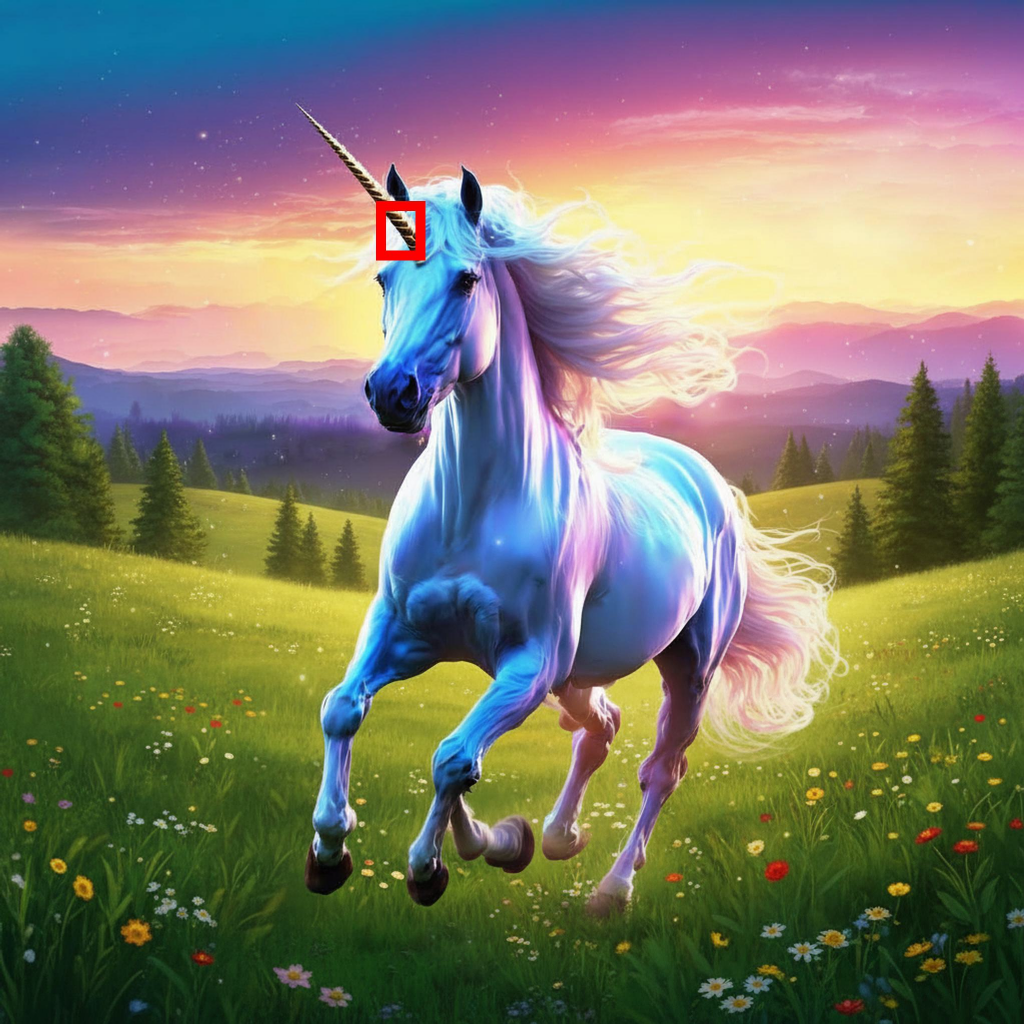} &
					\includegraphics[height=2.47cm]{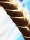} &
					\vdashline &
					\includegraphics[height=2.47cm]{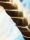}&
					\includegraphics[height=2.47cm]{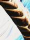} &
					\includegraphics[height=2.47cm]{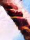} &
					\includegraphics[height=2.47cm]{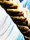} &
					\includegraphics[height=2.47cm]{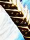}  \\
					\multicolumn{9}{c}{\textit{``Change the background to a green valley.''}}
				\end{tabular}
			\end{minipage} \\[3mm]
		\end{tabular}
	}
	\vspace{-2mm}
	\caption{\textbf{[Best visualized when magnified.]} Qualitative comparison of the \methodname~with diffusion-based super-resolution baselines~\cite{cheng2025effective, duan2025dit4sr, sun2024pisasr, dong2025tsd}. 
		First three rows show the results of 1K-editing, while the last three rows visualize 2K-editing.
		Here, we use the pretrained Stable Diffusion~\cite{rombach2022high} for~\methodname.
	}
	\label{fig:qual}
	\vspace{-4mm}
\end{figure*}

\subsection{Quantitative evaluation}
\label{subsec:exp_quant}

\paragraph{Baselines.}
A straightforward strategy for our task is to perform editing at a lower resolution and then apply super-resolution (SR) methods.
However, as discussed in the introduction, this approach fails to recover micro-scale textures because the fine-grained details of the source image are not conditioned during SR process. 
To verify that our claim is valid, we construct baselines by combining a state-of-the-art low-resolution editing method, Nano Banana~\cite{comanici2025gemini}, with various super-resolution approaches.
Specifically, we compare our method against (1) DiT-SR~\cite{cheng2025effective}, (2) DiT4SR~\cite{duan2025dit4sr}, (3) PiSA-SR~\cite{sun2024pisasr}, and (4) TSD-SR~\cite{dong2025tsd}.
We use a total of 400 (conditioning images, text instruction) pairs for evaluation of each editing scenario.

\begin{figure*}[t!]
	\centering
    \includegraphics[width=1.0\linewidth]{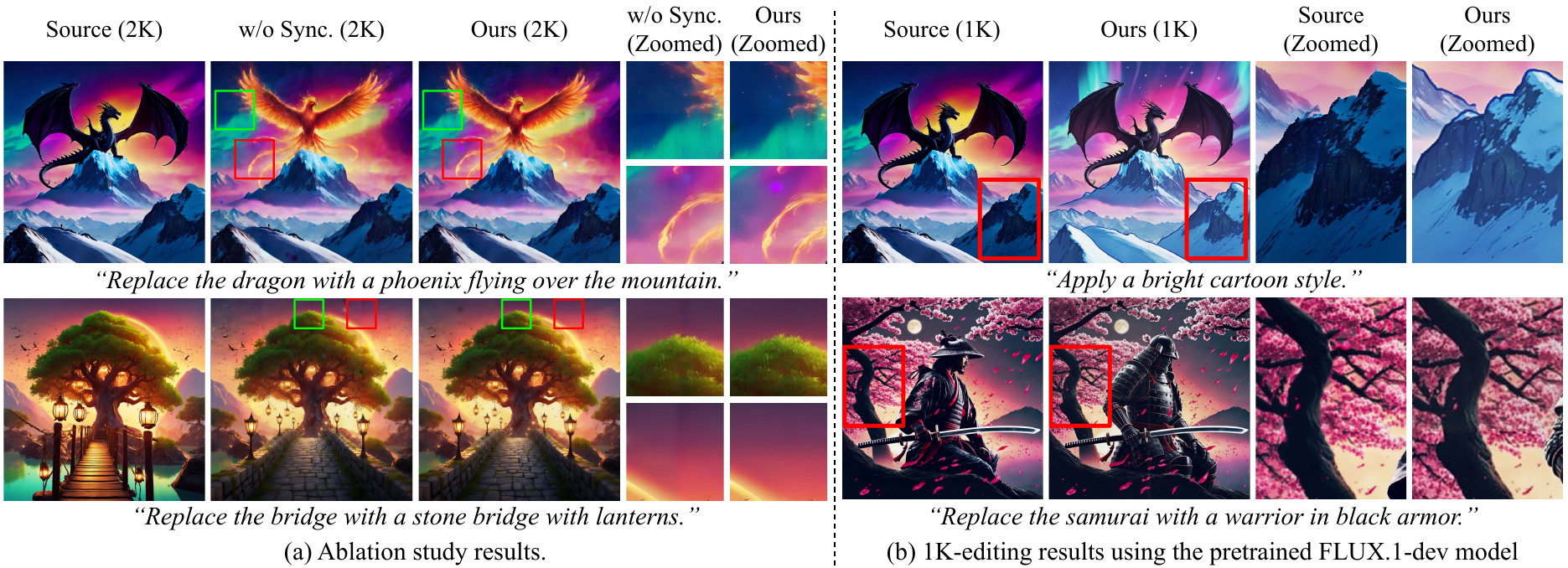}
    \vspace{-6mm}
	\caption{
    \textbf{[Best visualized when magnified.]} We visualize the effect of synchronization strategy in part (a), while we show the results of our method combined with the pretrained FLUX.1-dev~\cite{flux2024} model in part (b). 
            }  
	\vspace{-2mm}
\label{fig:qual_ablation}
\end{figure*}

\vspace{-10pt}
\paragraph{Metrics.}
We evaluate the results using five widely used metrics: MSE, SSIM, PSNR, LPIPS~\cite{zhang2018perceptual} and HaarPSI~\cite{reisenhofer2018haar}.
To measure MSE, SSIM, and PSNR, we compute their masked variants by applying region-specific masks to the source and target images, which are denoted as M-MSE, M-SSIM, and M-PSNR, respectively.
In particular, we use background masks for object change tasks and foreground masks for background modification tasks.
These masks are generated using LANG-SAM\footnote{https://github.com/luca-medeiros/lang-segment-anything}, which produces instruction-aware foreground and background segmentations.
We emphasize that the masked metrics allow us to access how well each method preserves the regions of the source image that are intended to remain unchanged during editing.
LPIPS quantifies the perceptual similarity between source and target images, which is widely used for image-to-image translation evaluation.
HaarPSI measures similarity in a wavelet-aware manner, enabling evaluation of how effectively fine-grained details are transferred from the source to the target image.

\vspace{-10pt}
\paragraph{Comparisons.}
We quantitatively compare our method with baselines under 1K-editing and 2K-editing scenarios.
Table~\ref{tab:quant} summarizes the evaluation results. 
For each column of the table, we \textbf{bold} and \underline{underline} the best and second-best result.
As shown, our method consistently outperforms all baselines, highlighting the effectiveness of the detail enhancement and synchronization mechanism.
In contrast, SR-based pipelines struggle to recover fine-grained details from the source image.
Moreover, the superior performance of \methodname~on masked-variant metrics indicates that it synthesizes the target image in a more source-aware manner, mitigating the inherent limitations of SR-based methods that exhibits source-unconditional generation.

\subsection{Qualitative results}
\label{subsec:exp_qual}

We present qualitative comparisons between our method and baselines using the pretrained Stable Diffusion~\cite{rombach2022high} in Figure~\ref{fig:qual}.
Here, the target image is expected to satisfy two criteria simultaneously:
(1) accurately inherit the fine-grained details from the zoomed source image and
(2) correctly reflect the semantics described in the text instruction.
Our method produces superior results, faithfully transferring fine-grained details of the source image to the target image while leveraging the desired semantics from the text instructions.
For example, in the 2nd row, the baselines struggle to adopt the ``watercolor'' style, whereas our method effectively renders the watercolor appearance while preserving the detail of the source image.
Similarly, in the 4th and 6th rows, our approach reliably carries over the texture-like details from the source, while the baselines exhibit noticeable color shifts or distorted textures.

\begin{figure}[t!]
	\centering
    \includegraphics[width=1.0\linewidth]{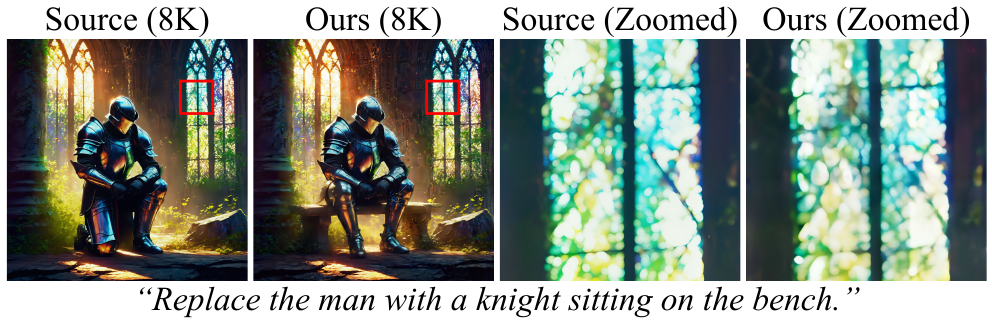}
	\vspace{-7mm}
	\caption{
    Qualitative results of \methodname~in 8K-editing scenario.
    }  
	\vspace{-4mm}
\label{fig:qual_8k}
\end{figure}

We visualize additional qualitative results in Figure~\ref{fig:teaser} and~\ref{fig:qual_ablation} (b).
Importantly, our method is not limited to Stable Diffusion architecture; 
it also generalizes to other backbones such as FLUX~\cite{flux2024}.
We also demonstrate the \emph{scalability} of the proposed method by visualizing 8K-editing results in Figure~\ref{fig:qual_8k}.
While existing image editing methods are either unable to handle 8K-editing scenario or require additional training, our method effectively performs editing regardless of the resolution without any additional tuning, enabled by the patch-wise detail transfer mechanism.

\subsection{Ablation study}
\label{subsec:exp_ablation}

We show the effectiveness of the proposed patch-wise synchronization strategy through an ablation study.
As illustrated in Figure~\ref{fig:qual_ablation} (a), the absence of synchronization leads to noticeable boundary artifacts, producing visible seams and inconsistencies along patch borders.
In contrast, synchronization enforces coherent denoising trajectories across neighboring patches, resulting in artifact-free transitions while preserving the local fidelity of each patch.

% !TEX root = ./../main.tex

\section{Conclusion}

In this work, we introduce the task of high-resolution image editing for the first time and propose a general framework to address it.
To fully inherit the strong prior of the pretrained image generation model, we divide the high-resolution source image into patches that match the model's native input resolution.
We then transfer fine-grained details from the source to the target image through a learnable transfer function that operates on intermediate features of the pretrained network.
To alleviate boundary artifacts between patches, we further propose a synchronization approach that eliminates the need for overlapping-view sampling used in prior synchronization methods, meaningfully reducing computational overhead.
Experiments across multiple resolution editing scenarios demonstrate that our method is capable of high-resolution image editing.

\section*{Acknowledgements}

This work was supported in part by NSF IIS2404180, and Institute of Information \& communications Technology Planning \& Evaluation (IITP) grants funded by the Korea government (MSIT) (No. 2022-0-00871, Development of AI Autonomy and Knowledge Enhancement for AI Agent Collaboration) and (No. RS-2022-00187238, Development of Large Korean Language Model Technology for Efficient Pre-training).
This work was also supported by Samsung Electronics Co., Ltd (IO250418-12669-01).

{
    \small
    \bibliographystyle{ieeenat_fullname}
    \bibliography{main}

@String(IJCV = {Int. J. Comput. Vis.})

@String(CVPR= {IEEE Conf. Comput. Vis. Pattern Recog.})

@String(ICCV= {Int. Conf. Comput. Vis.})

@String(ECCV= {Eur. Conf. Comput. Vis.})

@String(NIPS= {Adv. Neural Inform. Process. Syst.})

@String(ICLR = {Int. Conf. Learn. Represent.})

@String(AAAI = {AAAI})

@String(IJCV  = {IJCV})

@String(CVPR  = {CVPR})

@String(ICCV  = {ICCV})

@String(ECCV  = {ECCV})

@String(NIPS  = {NeurIPS})

@String(ICLR  = {ICLR})

@inproceedings{cheng2025effective,
  title={Effective diffusion transformer architecture for image super-resolution},
  author={Cheng, Kun and Yu, Lei and Tu, Zhijun and He, Xiao and Chen, Liyu and Guo, Yong and Zhu, Mingrui and Wang, Nannan and Gao, Xinbo and Hu, Jie},
  booktitle={AAAI},
  year={2025}
}

@inproceedings{duan2025dit4sr,
  title={DiT4SR: Taming Diffusion Transformer for Real-World Image Super-Resolution},
  author={Duan, Zheng-Peng and Zhang, Jiawei and Jin, Xin and Zhang, Ziheng and Xiong, Zheng and Zou, Dongqing and Ren, Jimmy and Guo, Chun-Le and Li, Chongyi},
  booktitle={ICCV},
  year={2025}
}

@inproceedings{sun2024pisasr,
  title={Pixel-level and Semantic-level Adjustable Super-resolution: A Dual-LoRA Approach},
  author={Sun, Lingchen and Wu, Rongyuan and Ma, Zhiyuan and Liu, Shuaizheng and Yi, Qiaosi and Zhang, Lei},
  booktitle={CVPR},
  year={2025}
}

@article{reisenhofer2018haar,
  title={A Haar wavelet-based perceptual similarity index for image quality assessment},
  author={Reisenhofer, Rafael and Bosse, Sebastian and Kutyniok, Gitta and Wiegand, Thomas},
  journal={Signal Processing: Image Communication},
  year={2018}
}

@article{comanici2025gemini,
  title={Gemini 2.5: Pushing the frontier with advanced reasoning, multimodality, long context, and next generation agentic capabilities},
  author={Comanici, Gheorghe and Bieber, Eric and Schaekermann, Mike and Pasupat, Ice and Sachdeva, Noveen and Dhillon, Inderjit and Blistein, Marcel and Ram, Ori and Zhang, Dan and Rosen, Evan and others},
  journal={arXiv:2507.06261},
  year={2025}
}

@inproceedings{qiu2025freescale,
  title={Freescale: Unleashing the resolution of diffusion models via tuning-free scale fusion},
  author={Qiu, Haonan and Zhang, Shiwei and Wei, Yujie and Chu, Ruihang and Yuan, Hangjie and Wang, Xiang and Zhang, Yingya and Liu, Ziwei},
  booktitle={ICCV},
  year={2025}
}

@inproceedings{mokady2023null,
  title={Null-text inversion for editing real images using guided diffusion models},
  author={Mokady, Ron and Hertz, Amir and Aberman, Kfir and Pritch, Yael and Cohen-Or, Daniel},
  booktitle={CVPR},
  year={2023}
}

@article{kingma2013auto,
  title={Auto-encoding variational bayes},
  author={Kingma, Diederik P and Welling, Max},
  journal={arXiv:1312.6114},
  year={2013}
}

@article{stein1981estimation,
	author = {Stein, Charles M},
	journal = {The annals of Statistics},
	title = "{Estimation of the Mean of a Multivariate Normal Distribution}",
	year = {1981}}

@article{song2020denoising,
  title={Denoising diffusion implicit models},
  author={Song, Jiaming and Meng, Chenlin and Ermon, Stefano},
  journal={ICLR},
  year={2021}
}

@article{ho2020denoising,
  title={Denoising diffusion probabilistic models},
  author={Ho, Jonathan and Jain, Ajay and Abbeel, Pieter},
  journal={NeurIPS},
  year={2020}
}

@inproceedings{sohl2015deep,
  title={Deep unsupervised learning using nonequilibrium thermodynamics},
  author={Sohl-Dickstein, Jascha and Weiss, Eric and Maheswaranathan, Niru and Ganguli, Surya},
  booktitle={ICML},
  year={2015},
}

@article{song2020score,
  title={Score-based generative modeling through stochastic differential equations},
  author={Song, Yang and Sohl-Dickstein, Jascha and Kingma, Diederik P and Kumar, Abhishek and Ermon, Stefano and Poole, Ben},
  journal={ICLR},
  year={2021}
}

@inproceedings{rombach2022high,
  title={High-resolution image synthesis with latent diffusion models},
  author={Rombach, Robin and Blattmann, Andreas and Lorenz, Dominik and Esser, Patrick and Ommer, Bj{\"o}rn},
  booktitle={CVPR},
  year={2022}
}

@misc{flux2024,
    author={Black Forest Labs},
    title={FLUX},
    year={2024},
    howpublished={\url{https://github.com/black-forest-labs/flux}},
}

@article{paszke2019pytorch,
  title={Pytorch: An imperative style, high-performance deep learning library},
  author={Paszke, Adam and Gross, Sam and Massa, Francisco and Lerer, Adam and Bradbury, James and Chanan, Gregory and Killeen, Trevor and Lin, Zeming and Gimelshein, Natalia and Antiga, Luca and others},
  journal={NeurIPS},
  year={2019}
}

@inproceedings{dong2025tsd,
  title={Tsd-sr: One-step diffusion with target score distillation for real-world image super-resolution},
  author={Dong, Linwei and Fan, Qingnan and Guo, Yihong and Wang, Zhonghao and Zhang, Qi and Chen, Jinwei and Luo, Yawei and Zou, Changqing},
  booktitle={CVPR},
  year={2025}
}

@inproceedings{zhang2018perceptual,
  title={The Unreasonable Effectiveness of Deep Features as a Perceptual Metric},
  author={Zhang, Richard and Isola, Phillip and Efros, Alexei A and Shechtman, Eli and Wang, Oliver},
  booktitle={CVPR},
  year={2018}
}

@article{vaswani2017attention,
  title={Attention is all you need},
  author={Vaswani, Ashish and Shazeer, Noam and Parmar, Niki and Uszkoreit, Jakob and Jones, Llion and Gomez, Aidan N and Kaiser, {\L}ukasz and Polosukhin, Illia},
  journal={NIPS},
  year={2017}
}

@article{wang2024exploiting,
  title={Exploiting diffusion prior for real-world image super-resolution},
  author={Wang, Jianyi and Yue, Zongsheng and Zhou, Shangchen and Chan, Kelvin CK and Loy, Chen Change},
  journal={IJCV},
  year={2024},
}

@inproceedings{wu2024seesr,
  title={Seesr: Towards semantics-aware real-world image super-resolution},
  author={Wu, Rongyuan and Yang, Tao and Sun, Lingchen and Zhang, Zhengqiang and Li, Shuai and Zhang, Lei},
  booktitle={CVPR},
  year={2024}
}

@article{ai2024dreamclear,
  title={DreamClear: High-Capacity Real-World Image Restoration with Privacy-Safe Dataset Curation},
  author={Ai, Yuang and Zhou, Xiaoqiang and Huang, Huaibo and Han, Xiaotian and Chen, Zhengyu and You, Quanzeng and Yang, Hongxia},
  journal={NeurIPS},
  year={2024}
}

@inproceedings{wang2024sinsr,
  title={Sinsr: diffusion-based image super-resolution in a single step},
  author={Wang, Yufei and Yang, Wenhan and Chen, Xinyuan and Wang, Yaohui and Guo, Lanqing and Chau, Lap-Pui and Liu, Ziwei and Qiao, Yu and Kot, Alex C and Wen, Bihan},
  booktitle={CVPR},
  year={2024}
}

@inproceedings{yu2024scaling,
  title={Scaling up to excellence: Practicing model scaling for photo-realistic image restoration in the wild},
  author={Yu, Fanghua and Gu, Jinjin and Li, Zheyuan and Hu, Jinfan and Kong, Xiangtao and Wang, Xintao and He, Jingwen and Qiao, Yu and Dong, Chao},
  booktitle={CVPR},
  year={2024}
}

@article{wu2024one,
  title={One-step effective diffusion network for real-world image super-resolution},
  author={Wu, Rongyuan and Sun, Lingchen and Ma, Zhiyuan and Zhang, Lei},
  journal={NeurIPS},
  year={2024}
}

@inproceedings{esser2024scaling,
  title={Scaling rectified flow transformers for high-resolution image synthesis},
  author={Esser, Patrick and Kulal, Sumith and Blattmann, Andreas and Entezari, Rahim and M{\"u}ller, Jonas and Saini, Harry and Levi, Yam and Lorenz, Dominik and Sauer, Axel and Boesel, Frederic and others},
  booktitle={ICML},
  year={2024}
}

@article{saharia2022photorealistic,
  title={Photorealistic text-to-image diffusion models with deep language understanding},
  author={Saharia, Chitwan and Chan, William and Saxena, Saurabh and Li, Lala and Whang, Jay and Denton, Emily L and Ghasemipour, Kamyar and Gontijo Lopes, Raphael and Karagol Ayan, Burcu and Salimans, Tim and others},
  journal={NeurIPS},
  year={2022}
}

@inproceedings{avrahami2022blended,
  title={Blended diffusion for text-driven editing of natural images},
  author={Avrahami, Omri and Lischinski, Dani and Fried, Ohad},
  booktitle={CVPR},
  year={2022}
}

@inproceedings{kim2022diffusionclip,
  title={Diffusionclip: Text-guided diffusion models for robust image manipulation},
  author={Kim, Gwanghyun and Kwon, Taesung and Ye, Jong Chul},
  booktitle={CVPR},
  year={2022}
}

@article{hertz2022prompt,
  title={Prompt-to-prompt image editing with cross attention control},
  author={Hertz, Amir and Mokady, Ron and Tenenbaum, Jay and Aberman, Kfir and Pritch, Yael and Cohen-Or, Daniel},
  journal={ICLR},
  year={2023}
}

@inproceedings{parmar2023zero,
  title={Zero-shot image-to-image translation},
  author={Parmar, Gaurav and Kumar Singh, Krishna and Zhang, Richard and Li, Yijun and Lu, Jingwan and Zhu, Jun-Yan},
  booktitle={SIGGRAPH},
  year={2023}
}

@inproceedings{tumanyan2023plug,
  title={Plug-and-play diffusion features for text-driven image-to-image translation},
  author={Tumanyan, Narek and Geyer, Michal and Bagon, Shai and Dekel, Tali},
  booktitle={CVPR},
  year={2023}
}

@inproceedings{cao2023masactrl,
  title={Masactrl: Tuning-free mutual self-attention control for consistent image synthesis and editing},
  author={Cao, Mingdeng and Wang, Xintao and Qi, Zhongang and Shan, Ying and Qie, Xiaohu and Zheng, Yinqiang},
  booktitle={ICCV},
  year={2023}
}

@article{lee2023conditional,
  title={Conditional score guidance for text-driven image-to-image translation},
  author={Lee, Hyunsoo and Kang, Minsoo and Han, Bohyung},
  journal={NeurIPS},
  year={2023}
}

@article{couairon2022diffedit,
  title={Diffedit: Diffusion-based semantic image editing with mask guidance},
  author={Couairon, Guillaume and Verbeek, Jakob and Schwenk, Holger and Cord, Matthieu},
  journal={ICLR},
  year={2023}
}

@inproceedings{lee2024diffusion,
  title={Diffusion-based image-to-image translation by noise correction via prompt interpolation},
  author={Lee, Junsung and Kang, Minsoo and Han, Bohyung},
  booktitle={ECCV},
  year={2024},
}

@inproceedings{brooks2023instructpix2pix,
  title={Instructpix2pix: Learning to follow image editing instructions},
  author={Brooks, Tim and Holynski, Aleksander and Efros, Alexei A},
  booktitle={CVPR},
  year={2023}
}

@article{epstein2023diffusion,
  title={Diffusion self-guidance for controllable image generation},
  author={Epstein, Dave and Jabri, Allan and Poole, Ben and Efros, Alexei and Holynski, Aleksander},
  journal={NeurIPS},
  year={2023}
}

@inproceedings{ronneberger2015u,
  title={U-net: Convolutional networks for biomedical image segmentation},
  author={Ronneberger, Olaf and Fischer, Philipp and Brox, Thomas},
  booktitle={International Conference on Medical image computing and computer-assisted intervention},
  year={2015},
}

@article{liu2025step1x-edit,
      title={Step1X-Edit: A Practical Framework for General Image Editing}, 
      author={Shiyu Liu and Yucheng Han and Peng Xing and Fukun Yin and Rui Wang and Wei Cheng and Jiaqi Liao and Yingming Wang and Honghao Fu and Chunrui Han and Guopeng Li and Yuang Peng and Quan Sun and Jingwei Wu and Yan Cai and Zheng Ge and Ranchen Ming and Lei Xia and Xianfang Zeng and Yibo Zhu and Binxing Jiao and Xiangyu Zhang and Gang Yu and Daxin Jiang},
      journal={arXiv:2504.17761},
      year={2025}
}

@inproceedings{peebles2023scalable,
  title={Scalable diffusion models with transformers},
  author={Peebles, William and Xie, Saining},
  booktitle={ICCV},
  year={2023}
}

@article{zhang2025context,
  title={In-context edit: Enabling instructional image editing with in-context generation in large scale diffusion transformer},
  author={Zhang, Zechuan and Xie, Ji and Lu, Yu and Yang, Zongxin and Yang, Yi},
  journal={NeurIPS},
  year={2025}
}

@article{hu2022lora,
  title={Lora: Low-rank adaptation of large language models.},
  author={Hu, Edward J and Shen, Yelong and Wallis, Phillip and Allen-Zhu, Zeyuan and Li, Yuanzhi and Wang, Shean and Wang, Lu and Chen, Weizhu and others},
  journal={ICLR},
  year={2022}
}

@article{zhu2025kv,
  title={KV-Edit: Training-Free Image Editing for Precise Background Preservation},
  author={Zhu, Tianrui and Zhang, Shiyi and Shao, Jiawei and Tang, Yansong},
  journal={ICCV},
  year={2025}
}

@article{xiao2023efficient,
  title={Efficient streaming language models with attention sinks},
  author={Xiao, Guangxuan and Tian, Yuandong and Chen, Beidi and Han, Song and Lewis, Mike},
  journal={ICLR},
  year={2024}
}

@article{podell2023sdxl,
  title={Sdxl: Improving latent diffusion models for high-resolution image synthesis},
  author={Podell, Dustin and English, Zion and Lacey, Kyle and Blattmann, Andreas and Dockhorn, Tim and M{\"u}ller, Jonas and Penna, Joe and Rombach, Robin},
  journal={ICLR},
  year={2024}
}

@inproceedings{huang2024fouriscale,
  title={Fouriscale: A frequency perspective on training-free high-resolution image synthesis},
  author={Huang, Linjiang and Fang, Rongyao and Zhang, Aiping and Song, Guanglu and Liu, Si and Liu, Yu and Li, Hongsheng},
  booktitle={ECCV},
  year={2024}
}

@inproceedings{du2024demofusion,
  title={Demofusion: Democratising high-resolution image generation with no \$\$\$},
  author={Du, Ruoyi and Chang, Dongliang and Hospedales, Timothy and Song, Yi-Zhe and Ma, Zhanyu},
  booktitle={CVPR},
  year={2024}
}

@inproceedings{he2023scalecrafter,
  title={Scalecrafter: Tuning-free higher-resolution visual generation with diffusion models},
  author={He, Yingqing and Yang, Shaoshu and Chen, Haoxin and Cun, Xiaodong and Xia, Menghan and Zhang, Yong and Wang, Xintao and He, Ran and Chen, Qifeng and Shan, Ying},
  booktitle={ICLR},
  year={2024}
}

@article{hu2025ultragen,
  title={UltraGen: High-Resolution Video Generation with Hierarchical Attention},
  author={Hu, Teng and Zhang, Jiangning and Su, Zihan and Yi, Ran},
  journal={AAAI},
  year={2026}
}

@inproceedings{blattmann2023align,
  title={Align your latents: High-resolution video synthesis with latent diffusion models},
  author={Blattmann, Andreas and Rombach, Robin and Ling, Huan and Dockhorn, Tim and Kim, Seung Wook and Fidler, Sanja and Kreis, Karsten},
  booktitle={CVPR},
  year={2023}
}

@inproceedings{zhang2025diffusion,
  title={Diffusion-4k: Ultra-high-resolution image synthesis with latent diffusion models},
  author={Zhang, Jinjin and Huang, Qiuyu and Liu, Junjie and Guo, Xiefan and Huang, Di},
  booktitle={CVPR},
  year={2025}
}

@article{niu2025mineru2,
  title={Mineru2. 5: A decoupled vision-language model for efficient high-resolution document parsing},
  author={Niu, Junbo and Liu, Zheng and Gu, Zhuangcheng and Wang, Bin and Ouyang, Linke and Zhao, Zhiyuan and Chu, Tao and He, Tianyao and Wu, Fan and Zhang, Qintong and others},
  journal={arXiv:2509.22186},
  year={2025}
}

@article{chen2025dc,
  title={DC-VideoGen: Efficient Video Generation with Deep Compression Video Autoencoder},
  author={Chen, Junyu and He, Wenkun and Gu, Yuchao and Zhao, Yuyang and Yu, Jincheng and Chen, Junsong and Zou, Dongyun and Lin, Yujun and Zhang, Zhekai and Li, Muyang and others},
  journal={arXiv:2509.25182},
  year={2025}
}

@inproceedings{wu2025drivescape,
  title={DriveScape: High-Resolution Driving Video Generation by Multi-View Feature Fusion},
  author={Wu, Wei and Guo, Xi and Tang, Weixuan and Huang, Tingxuan and Wang, Chiyu and Ding, Chenjing},
  booktitle={CVPR},
  year={2025}
}

@misc{lai2025hunyuan3d25highfidelity3d,
      title={Hunyuan3D 2.5: Towards High-Fidelity 3D Assets Generation with Ultimate Details}, 
      author={Tencent Hunyuan3D Team},
      year={2025},
      eprint={2506.16504},
      archivePrefix={arXiv},
      primaryClass={cs.CV},
      url={https://arxiv.org/abs/2506.16504}, 
}

@misc{hunyuan3d22025tencent,
    title={Hunyuan3D 2.0: Scaling Diffusion Models for High Resolution Textured 3D Assets Generation},
    author={Tencent Hunyuan3D Team},
    year={2025},
    eprint={2501.12202},
    archivePrefix={arXiv},
    primaryClass={cs.CV}
}

@inproceedings{radford2021learning,
  title={Learning transferable visual models from natural language supervision},
  author={Radford, Alec and Kim, Jong Wook and Hallacy, Chris and Ramesh, Aditya and Goh, Gabriel and Agarwal, Sandhini and Sastry, Girish and Askell, Amanda and Mishkin, Pamela and Clark, Jack and others},
  booktitle={ICML},
  year={2021},
}

@inproceedings{he2016deep,
  title={Deep residual learning for image recognition},
  author={He, Kaiming and Zhang, Xiangyu and Ren, Shaoqing and Sun, Jian},
  booktitle={CVPR},
  year={2016}
}

@article{lipman2022flow,
  title={Flow matching for generative modeling},
  author={Lipman, Yaron and Chen, Ricky TQ and Ben-Hamu, Heli and Nickel, Maximilian and Le, Matt},
  journal={ICLR},
  year={2023}
}

@article{liu2022flow,
  title={Flow straight and fast: Learning to generate and transfer data with rectified flow},
  author={Liu, Xingchao and Gong, Chengyue and Liu, Qiang},
  journal={ICLR},
  year={2023}
}

@inproceedings{lee2025diffusion,
  title={Diffusion-based conditional image editing through optimized inference with guidance},
  author={Lee, Hyunsoo and Kang, Minsoo and Han, Bohyung},
  booktitle={WACV},
  year={2025},
}

@article{kang2025icm,
  title={ICM-SR: Image-Conditioned Manifold Regularization for Image Super-Resoultion},
  author={Kang, Junoh and Ryou, Donghun and Han, Bohyung},
  journal={arXiv:2511.22048},
  year={2025}
}
}

% \clearpage
% \appendix 
% !TEX root = ./../main.tex

\clearpage
\setcounter{equation}{14}
\setcounter{table}{1}
\setcounter{figure}{6}
\maketitlesupplementary

\appendix
\section{Pseudo-code of~\methodname}

We provide an abstract overview of the proposed method by outlining its core procedure in Algorithm~\ref{alg:overview}.

% !TEX root = ./../main.tex

\begin{algorithm}[h!]
	\caption{\methodname}
	\label{alg:overview}
	\begin{algorithmic}[1]
		
		\State \textbf{Inputs:} Conditioning images $\imghs$, $\imgls$, $\imglr$, hyperparameter $\tau$
		
        \noindent // 1. Forward process
		\For{$i \gets 1, \cdots , NM$}
        \State Get $\{\srch_t[i] \}_{t=0}^{T}, \{\srcl_t[i] \}_{t=0}^{T}, \{\refl_t [i]\}_{t=0}^{T}$
        \State $\mathbf{\tilde{x}}_T[i] \gets \srcl_T[i]$, $\mathbf{\tilde{y}}_T[i] \gets \refl_T[i]$
        \EndFor

        \noindent // 2. Transfer function optimization
        \For{$t\gets T, \cdots, 1$, \; $i \gets 1, \cdots, NM$}
            \If {$t \leq \tau$} 
                \State Optimize $\phi_{\theta}$ using Eq. (6) 
                \State $\Delta \mathbf{h}_t[i] \gets \phi_{\theta}(\mathbf{h}_t[i], t)$
            \Else
                \State $\Delta \mathbf{h}_t[i] \gets \mathbf{0}$
            \EndIf
            \State Get $\mathbf{\tilde{x}}_{t-1}[i]$ using Eq. (7) 
        \EndFor

        \noindent // 3. Detail injection with synchronization
        \For{$t\gets T, \cdots, 1$}
             \State Get $\mathbf{\tilde{y}}_{t-1}[i]$ using Eq. (9)  for $1 \leq i \leq NM$
            \If {$t \leq \tau$}
                \State Get $\mathbf{\tilde{y}}_{t}^{\mathrm{rsp}}[i]$ using Eq. (13) for $1 \leq i \leq NM$
                \State Get $\mathbf{\tilde{y}}_{t-1}[i]$ using Eq. (14)  for $1 \leq i \leq NM$
            \EndIf
        \EndFor

		\State $\imghr \gets \mathrm{Decode}(\mathrm{Composite}( \{\mathbf{\tilde{y}}_0[i] \}_{i=1}^{NM}) )$
		\State {\bf Output:} Target image $\imghr$
	\end{algorithmic}
\end{algorithm}

\section{Detailed explanation on synchronization}
\label{subsec:supp_sync}

To synchronize adjacent patches, we introduce an auxiliary latent $\mathbf{\tilde{A}}_t [i, i+1]$, constructed by spatially blending the bottom half of $\mathbf{\tilde{y}}^{\mathrm{rsp}}_t[i]$ and the upper half of $\mathbf{\tilde{y}}^{\mathrm{rsp}}_t[i+1]$ at timestep $t$.
Since each patch is denoised independently, their latent trajectories may diverge, often producing visible discontinuities along their boundary.
However, $\mathbf{\tilde{A}}_t[i, i+1]$ contains the boundary region, and therefore its Tweedie estimate may naturally captures a smoother transition between the two patches.
By blending the Tweedie estimate of an auxiliary patch with those of the original patches, the boundary between neighboring patches becomes more coherent, reducing artifacts and enhancing spatial continuity.

For the spatial blending, we first apply  Eq. (2) to obtain the original patches' Tweedie estimates:
\begin{align}
    \hat{\mathbf{x}}_{t \rightarrow 0}^{\mathrm{aux}}[i, i+1] &= \hat{\mathbf{x}}_0 (\mathbf{\tilde{A}}_t[i, i+1], t) \\
    \hat{\mathbf{x}}_{t \rightarrow 0}[i] &= \hat{\mathbf{x}}_0 (\mathbf{\tilde{y}}^{\mathrm{rsp}}_t[i], t).
\end{align}
{Then we define $L(\hat{\mathbf{x}}_{t \rightarrow 0}[i], \hat{\mathbf{x}}_{t \rightarrow 0}^{\mathrm{aux}}[i, i+1])$, a blended Tweedie estimate that is used to swap the $\hat{\mathbf{x}}_0$ term in Eq. (1) during the reverse process of $\mathbf{\tilde{y}}^{\mathrm{rsp}}_t[i]$.}
Let $\mathcal{T}_1$ be the vertical translation operator defined in $[0, H_p] \times [0, W_p] $ as follows:
\begin{equation}
    \mathcal{T}_1f(u, v) = 
    \begin{cases}
        0, & 0 \leq v < H_p / 2, \\
        f(u, v-H_p / 2), & H_p / 2 \leq v \leq H_p,
    \end{cases}
\end{equation}
where $W_{{p}}$ and $H_p$ denote the patch width and height, respectively.
Thus, $\mathcal{T}_1\hat{\mathbf{x}}_{t \rightarrow 0}^{\mathrm{aux}}[i, i+1]$ repositions $\hat{\mathbf{x}}_{t \rightarrow 0}^{\mathrm{aux}}[i, i+1]$ so that its top edge aligns with the  midpoint of $\hat{\mathbf{x}}_{t \rightarrow 0}[i]$.
Then we define a spatial weight mask $\mathbf{M}_1(u, v, t)$ for smooth transition across the overlap:
\begin{equation}
    \mathbf{M}_1(u, v, t) = 
    \begin{cases}
    0, & 0 \leq v < H_p/2, \\ 
    \frac{v - H_p/2}{H_p/2} \cdot \left(1 - \frac{t}{\tau} \right), & H_p / 2 \leq v \leq H_p.
    \end{cases}
\end{equation}
The blended Tweedie estimate is then computed as
\begin{align}
& L(\hat{\mathbf{x}}_{t \rightarrow 0}[i], \hat{\mathbf{x}}_{t \rightarrow 0}^{\mathrm{aux}}[i, i+1])  \\ &= (1 - \mathbf{M}_1 ) \odot \hat{\mathbf{x}}_{t \rightarrow 0}[i] + \mathbf{M}_1 \odot \mathcal{T}_1 \hat{\mathbf{x}}_{t \rightarrow 0}^{\mathrm{aux}}[i, i+1]. \nonumber 
\label{eq:blended_tweedie_estimate}
\end{align}
For the rest of the paragraph, we detail the case of $\mathbf{\tilde{y}}^{\mathrm{rsp}}_t[i+1]$.
Let $\mathcal{T}_2$ and $\mathbf{M}_2 (u, v, t)$ defined in the vertically-flipped manner compared to $\mathcal{T}_1$ and $\mathbf{M}_1(u, v, t)$:
\begin{equation}
    \mathcal{T}_2f(u, v) = 
    \begin{cases}
        f(u, v+H_{{p}} / 2), & 0 \leq v \leq H_p / 2 \\
        0, & H_p / 2 < v < H_p, 
    \end{cases}
\end{equation}
\begin{equation}
    \mathbf{M}_2(u, v, t) = 
    \begin{cases}
    \frac{H_p/2 - v }{H_p/2} \cdot \left(1 - \frac{t}{\tau} \right), & 0\leq v \leq H_p/2 \\
    0, & H_p/2 < v < H_p.
    \end{cases}
\end{equation}
Subsequently, the blended Tweedie estimate is calculated as:
\begin{align}
& L(\hat{\mathbf{x}}_{t \rightarrow 0}[i+1], \hat{\mathbf{x}}_{t \rightarrow 0}^{\mathrm{aux}}[i, i+1])   \\ &= (1 - \mathbf{M}_2 ) \odot \hat{\mathbf{x}}_{t \rightarrow 0}[i+1] + \mathbf{M}_2 \odot \mathcal{T}_2 \hat{\mathbf{x}}_{t \rightarrow 0}^{\mathrm{aux}}[i, i+1]. \nonumber
\end{align}

\section{Implementation details}
\label{subsec:supp_imple_details}

\paragraph{Latent decoding.}

After denoising $N \times M$ latent patches, we first merge them into a single latent tensor.
The merged latent is then passed through the pretrained VAE decoder~\cite{kingma2013auto} to obtain the final high-resolution image $\imghr$.

\paragraph{Adaptation to Stable Diffusion.}

For Stable Diffusion~\cite{rombach2022high} v2.1-base, the upsampling path of the U-Net~\cite{ronneberger2015u} consists of four upsampling blocks, each composed of three sub-blocks.
We insert a $1 \times 1$ convolution layer into the last sub-block of each upsampling block, resulting in four additional $1 \times 1$ convolution layers in total.

\paragraph{Adaptation to FLUX.}

For FLUX.1-dev~\cite{flux2024}, we attach the detail enhancement module to the linear layer located immediately after the last single-stream block, which is followed by a Layer Normalization operation.
Since FLUX is based on flow matching~\cite{lipman2022flow, liu2022flow}, we apply \methodname~with several additional modifications.
The forward and reverse processes follow deterministic ODE dynamics, defined as:
\begin{equation}
    \mathbf{x}_{t+1}^{\mathrm{fwd}}  =  \mathbf{x}_{t}^{\mathrm{fwd}} + (\sigma_{t+1} - \sigma_t) \cdot  \mathbf{v}_{\theta}(\mathbf{x}_{t}^{\mathrm{fwd}} , t),
\end{equation}
\begin{equation}
    \mathbf{x}_{t-1}^{\mathrm{rev}}  =  \mathbf{x}_{t}^{\mathrm{rev}} + (\sigma_{t-1} - \sigma_t) \cdot  \mathbf{v}_{\theta}(\mathbf{x}_{t}^{\mathrm{rev}} , t),
\end{equation}
where $\mathbf{v}_{\theta}(\cdot, \cdot)$ denotes the pretrained vector field prediction network.
The clean data sample is estimated as:
\begin{equation}
    \hat{\mathbf{x}}_0(\mathbf{x}_t, t) = \mathbf{x}_t - \sigma_t \mathbf{v}_{\theta}(\mathbf{x}_t, t),
\end{equation}
which corresponds to the Tweedie estimate in diffusion models.
In addition, the reverse process using the blended latent differs from that of diffusion models.
We calculate the blended latent $L(\hat{\mathbf{x}}_{t \rightarrow 0}[i], \hat{\mathbf{x}}_{t \rightarrow 0}^{\mathrm{aux}}[i, i+1])$ using the same blending operation described in Sec.~4.5.1 of the main paper, which corresponds to the blended Tweedie estimate in diffusion models.
We then compute the modified vector field using the blended latent as follows:
\begin{equation}
    \mathbf{v}'(t)[i] = \frac{\mathbf{\tilde{y}}_t^{\mathrm{rsp}}[i] - L(\hat{\mathbf{x}}_{t \rightarrow 0}[i], \hat{\mathbf{x}}_{t \rightarrow 0}^{\mathrm{aux}}[i, i+1])}{\sigma_t},
\end{equation}
Finally, we perform the reverse process using the modified vector field:
\begin{equation}
   \mathbf{\tilde{y}}_{t-1}[i] = \mathbf{\tilde{y}}_t^{\mathrm{rsp}}[i] + (\sigma_{t-1} - \sigma_t) \cdot  \mathbf{v}'(t)[i].
\end{equation}
In practice, to achieve accurate reconstruction, we cache the output of each attention layer in the forward process and reuse it in the reverse process.

\section{Additional ablation studies}
\paragraph{Impact of hyperparameter $\tau$.} 

As described in Algorithm~\ref{alg:overview}, ScaleEdit utilizes a hyperparameter $\tau$. To investigate its effect, we conduct an ablation study by varying $\tau$ in ${15, 25, 35}$ and evaluate the results using the same metrics as the main experiment on a set of 60 (image, instruction) pairs. As shown in Table~\ref{tab:ablation_tau}, the default setting $\tau = 15$ achieves the best performance across all configurations.

\begin{table}[!h]
{
\renewcommand{\arraystretch}{1.2}
\setlength{\tabcolsep}{6pt}
\centering
\caption{Ablation study on $\tau$ values.}
\vspace{-3mm}
\scalebox{0.8}{
\begin{tabular}{cccccc}
	\toprule
    $\tau$ & HaarPSI $\uparrow$ & M-MSE $\downarrow$ & M-SSIM $\uparrow$ & M-PSNR $\uparrow$ & LPIPS $\downarrow$ \\
	\hline
	$15$ & 0.335 & 0.042 & 0.573 & 17.430 & 0.472 \\
	$25$ & 0.326 & 0.043 & 0.561 & 17.097 & 0.471 \\
	$35$ & 0.308 & 0.044 & 0.537 & 16.308 & 0.496 \\
	\bottomrule
\end{tabular}
}
\vspace{-5mm}
\label{tab:ablation_tau}
\par
}
\end{table}

\paragraph{Ablation on the design choice of transfer function.}

We first analyze the effect of layers by varying the index of the sub-block used within each block of the U-Net~\cite{ronneberger2015u} that are used for the detail transfer module.
Specifically, for each stage (down/middle/up), we select $N$-th sub-block from all blocks in that stage.
Variant of $\phi_{\theta}$ is also evaluated by replacing the convolution with constant mapping.
Evaluation is conducted using a total of 60 (image, instruction) pairs.
Tab.~\ref{tab:layer_selection} shows our design (Up \#3) performs best; 
while maintaining the overall perceptual similarity (measured by LPIPS~\cite{zhang2018perceptual}), it also preserves the intended editing effects (quantified by CLIP-Sim~\cite{radford2021learning} between target image and target prompt).
Alternatives still outperform most baselines but fall short.
Since the last (3rd) sub-blocks in up stage operate at high-resolution, they are most effective at transferring fine-grained details needed for high-resolution generation.

\vspace{-2mm}
\begin{table}[h!]
\centering
\caption{Ablation study on design of detail transfer module.}
\vspace{-3mm}
\small
\setlength{\tabcolsep}{2pt}
\renewcommand{\arraystretch}{1.2}
\resizebox{1.0\columnwidth}{!}{
\begin{tabular}{lccccccc}
\hline
Method & Down \#1 & Down \#2 & Middle & Up \#1  & Up \#2 & \textbf{Up \#3} & Constant \\
\hline
LPIPS $\downarrow$  & 0.1718 & 0.1746 & 0.1913 & \textbf{0.1648}  & 0.1663 & \textbf{0.1648} & 0.1781 \\
CLIP-Sim. $\uparrow$ &0.2676 &0.2677 & 0.2674 & 0.2681 & 0.2681 & \textbf{0.2684} & 0.2683 \\
\hline
\end{tabular}
}
\vspace{-5mm}
\label{tab:layer_selection}
\end{table}

\paragraph{Ablation on synchronization strategy.}

In Figure~\ref{fig:supp_sync}, we illustrate the effect of the proposed synchronization method. 
As shown, naive patch sampling without synchronization brings noticeable edge artifacts along patch boundaries, whereas our method effectively mitigates these artifacts.

\section{Receptive field of generative models}
In this work, we leverage a low-resolution image generation model to perform high-resolution image editing in a patch-wise manner. 
Although FreeScale~\cite{qiu2025freescale} modifies internal components of a low‑resolution diffusion model (\textit{e.g.} via dilated convolutions) to generate high‑resolution outputs, such architectural changes do not guarantee a strong high‑resolution image prior, as the model is not trained on high‑resolution image datasets.

In contrast, our patch-wise approach is well-suited for extracting detailed information.
Each patch matches the resolution of the low-resolution model’s receptive field (\textit{e.g.} $512 \times 512$ for the pretrained Stable Diffusion~\cite{rombach2022high}), enabling faithful detail reconstruction without modifying the underlying generator.
The main challenge in the modified framework lies in generating details that the model was not originally trained to produce.
For these reasons, we adopt a patch-wise strategy grounded in low-resolution image priors instead of modifying diffusion architectures to synthesize high-resolution content directly.

\section{Additional results}

\begin{figure}[h!]
	\centering
    \includegraphics[width=1.0\linewidth]{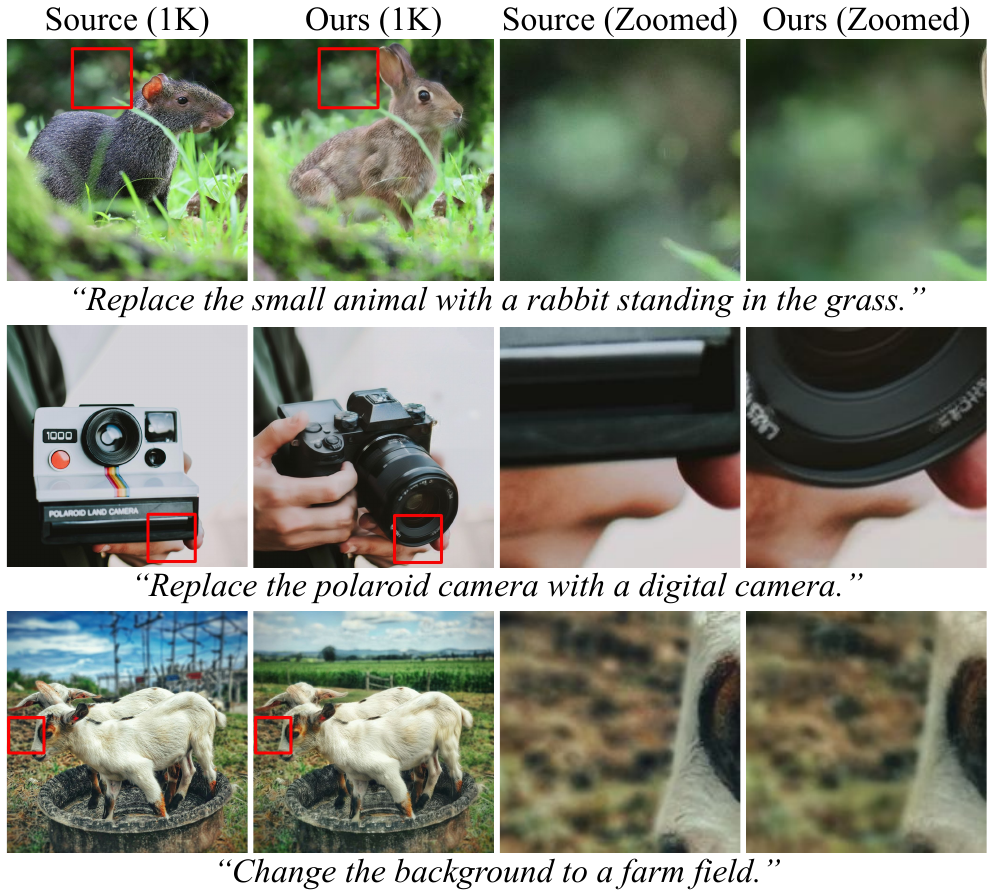}
	\vspace{-7mm}
	\caption{
    \textbf{[Best visualized when magnified.]} {Qualitative results of real image editing in 1K resolution.
    }}  
	\vspace{-6mm}
\label{fig:real_editing}
\end{figure}

To demonstrate the performance of \methodname~in real scenarios, we conduct additional 1K-editing experiments using a total of 285 (image, text instruction) pairs.
We sample high-resolution real images from the publicly accessible Internet source\footnote{https://www.pexels.com/}.
Then we follow the procedure described in Sec.~5.2 of the main paper for experiments. 
Figure~\ref{fig:real_editing} and Table~\ref{tab:real_editing} shows the result of real image editing.
As shown, our method demonstrates strong performance.

\begin{figure}[h!]
	\centering
	\includegraphics[width=1.0\linewidth]{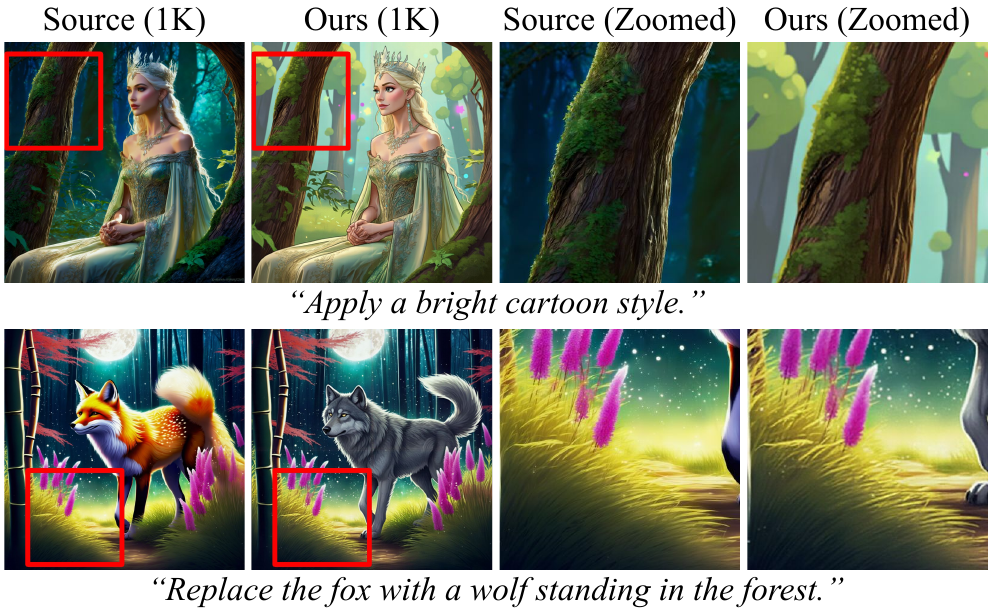}
	\vspace{-3mm}
	\caption{\textbf{[Best visualized when magnified.]} 
    Qualitative results of our method combined with the pretrained FLUX.1-dev~\cite{flux2024} model.}
	\vspace{-4mm}
\label{fig:supp_flux}
\end{figure}

\begin{figure*}[t!]
	\centering
	\includegraphics[width=0.9\linewidth]{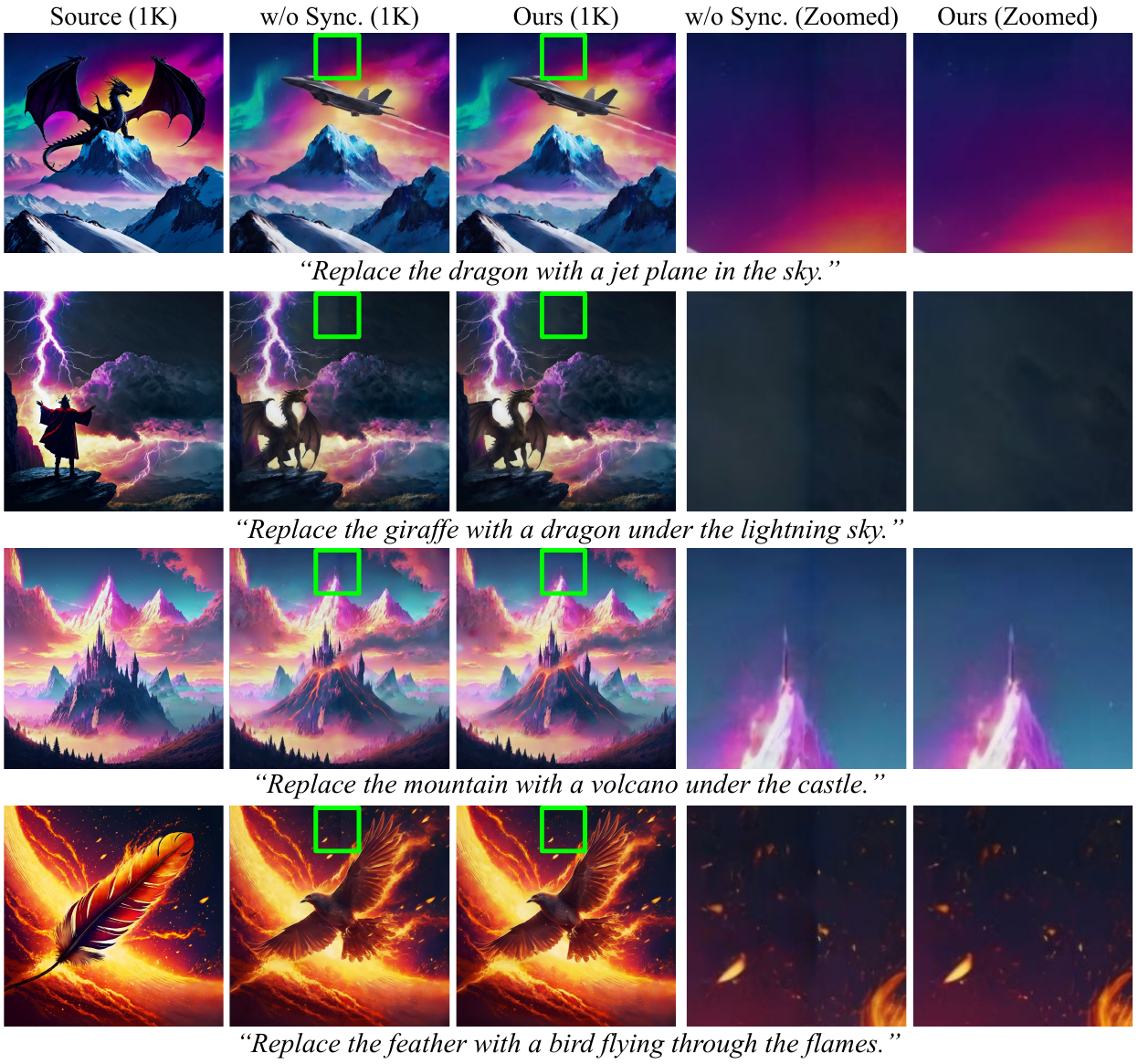}
	\caption{\textbf{[Best visualized when magnified.]} Effectiveness of the proposed synchronization strategy. 
    While images sampled without synchronization incorporates a significant artifact on the patch boundaries, our method effectively alleviates the edge artifacts.}
\label{fig:supp_sync}
\end{figure*}

\begin{figure*}[t!]
	\centering
	\includegraphics[width=1.0\linewidth]{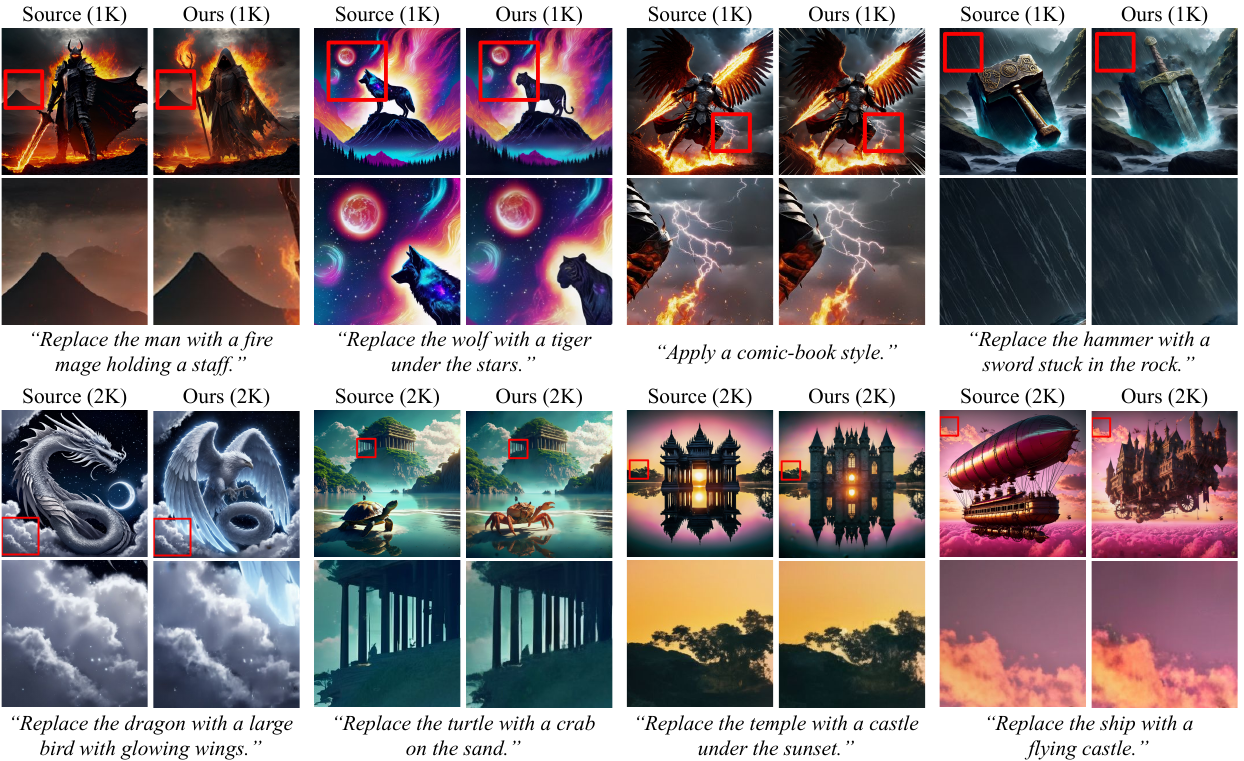}
	\caption{\textbf{[Best visualized when magnified.]}
    We visualize the additional results of 1K-editing and 2K-editing. 
    By conditioning on low-resolution reference images, our method is successfully synthesizes high-resolution edited images.
    }
\label{fig:supp_ours}
\end{figure*}

We also show the additional results of \methodname~on synthetic images in Figure~\ref{fig:supp_flux} and ~\ref{fig:supp_ours}.
Figure~\ref{fig:supp_flux} shows that our method is even applicable to transformer-based~\cite{vaswani2017attention} FLUX model~\cite{flux2024}, demonstrating the robustness and generalizability of our method across backbone architectures.
Then, we show the additional 1K- and 2K-editing results obtained with the pretrained Stable Diffusion~\cite{rombach2022high} in Figure~\ref{fig:supp_ours}.
We emphasize that the proposed method effectively transfers the fine-grained details of the source image into the target image.

\vspace{-2mm}

\begin{table}[h!]
{
\renewcommand{\arraystretch}{1.2}
\setlength{\tabcolsep}{4pt}
\centering
\caption{Quantitative evaluation under 1K-editing scenario with real images using the pretrained Stable Diffusion~\cite{rombach2022high}.}
\vspace{-2mm}
\scalebox{0.72}{
\begin{tabular}{lccccc}
	\toprule
    Method & HaarPSI $\uparrow$ & M-MSE $\downarrow$ & M-SSIM $\uparrow$ & M-PSNR $\uparrow$ & LPIPS $\downarrow$ \\
	\hline
	DiT-SR~\cite{cheng2025effective} & \underline{0.347} & 0.068 & 0.692 & 22.407 & 0.148  \\
	DiT4SR~\cite{duan2025dit4sr} & 0.345 & 0.068 & 0.706 & 22.377 & 0.141 \\
	PiSA-SR~\cite{sun2024pisasr} & 0.346 & \underline{0.067} & \underline{0.719} & \underline{22.658} & \underline{0.137} \\
	TSD-SR~\cite{dong2025tsd} & \underline{0.347} & 0.071 & 0.693 & 22.127 & 0.139 \\
	\textbf{\methodname~(Ours)} & \textbf{0.375} & \textbf{0.065} & \textbf{0.781} & \textbf{23.270} & \textbf{0.134} \\
	\bottomrule
\end{tabular}
}
\label{tab:real_editing}
\par
}
\end{table}

\section{Discussion on computational cost}

We evaluate the computational efficiency of our method against baselines in terms of runtime and GPU memory usage using a single NVIDIA A6000 GPU on 1K-editing scenario.
While baselines in half-precision typically require 0.74–88.06 seconds and 8.40–36.00 GB of VRAM, our method currently operates in full-precision to ensure numerical stability during optimization, requiring 20.29 GB. 
Although our method currently has a higher overhead, we anticipate that further refinement of the implementation will enable comparable efficiency in half-precision without compromising performance.

We also note that our framework is compatible with Null-text Inversion (NTI)~\cite{mokady2023null} for accurate reconstruction. 
The runtime without and with NTI are 234.77 and 635.51 seconds, respectively; 
we clarify that latter case's overhead is mainly due to NTI’s iterative optimization rather than our core components. 
Ultimately, these results represent a justifiable trade-off for the state-of-the-art performance our method achieves in Table 1 and Figure 4, consistently outperforming baselines in high-resolution image editing.
Note that for the evaluation results reported in the main paper, all methods were executed in full-precision whenever possible, unless this led to out-of-memory errors.

\section{Limitations}

Our method may produce suboptimal results due to the limited performance of the pretrained generative models~\cite{rombach2022high, flux2024}. 
Furthermore, since the proposed method relies on the low-resolution reference image $\imglr$ generated by existing low-resolution image-to-image translation methods, some artifacts introduced by these editing methods can propagate to the final high-resolution output $\imghr$, potentially leading to visible artifacts.

\section{Societal impacts}

The proposed method may generate some harmful images due to the imperfections of the underlying pretrained generative models~\cite{rombach2022high, flux2024}.
Rare cases of undesired or inappropriate results may arise, especially when the generative prior itself produces such outputs under challenging conditions.

\end{document}